\documentclass{article}

\usepackage[preprint]{neurips_2024}

\usepackage[utf8]{inputenc} 
\usepackage[T1]{fontenc}    
\usepackage{hyperref}       
\usepackage{url}            
\usepackage{booktabs}       
\usepackage{amsfonts}       
\usepackage{nicefrac}       
\usepackage{microtype}      
\usepackage{xcolor}         

\usepackage{amsmath}
\usepackage{amssymb}
\usepackage{mathtools}
\usepackage{amsthm}
\usepackage{bigints}
\usepackage{caption}
\usepackage{subcaption}
\usepackage{graphicx}
\usepackage{multirow}
\usepackage{bbm}
\usepackage{relsize}
\usepackage[scaled=0.8]{beramono}

\theoremstyle{plain}
\newtheorem{theorem}{Theorem}[section]
\numberwithin{equation}{theorem}

\theoremstyle{definition}
\newtheorem{definition}[theorem]{Definition}

\theoremstyle{remark}

\title{Transformers are Expressive, But Are They Expressive Enough for Regression?}

\author{%
  Swaroop Nath, Harshad Khadilkar and Pushpak Bhattacharyya \\
  Computer Science and Engineering, IIT Bombay\\
  India \\
  \texttt{\{swaroopnath, harshadk, pb\}@cse.iitb.ac.in} \\
}

\begin{document}

\maketitle

\begin{abstract}
  Transformers have become pivotal in Natural Language Processing, demonstrating remarkable success in applications like Machine Translation and Summarization. Given their widespread adoption, several works have attempted to analyze the expressivity of Transformers. Expressivity of a neural network is the class of functions it can approximate. A neural network is fully expressive if it can act as a universal function approximator. We attempt to analyze the same for Transformers. Contrary to existing claims, our findings reveal that Transformers struggle to reliably approximate smooth functions, relying on piecewise constant approximations with sizable intervals. The central question emerges as: ``\textit{Are Transformers truly Universal Function Approximators}?'' To address this, we conduct a thorough investigation, providing theoretical insights and supporting evidence through experiments. Theoretically, we prove that \textit{Transformer Encoders} cannot approximate smooth functions. Experimentally, we complement our theory and show that the \textit{full Transformer architecture} cannot approximate smooth functions. By shedding light on these challenges, we advocate a refined understanding of Transformers' capabilities. Code Link: \href{https://github.com/swaroop-nath/transformer-expressivity}{\tt github.com/transformer-expressivity}.
\end{abstract}

\section{Introduction}\label{sec:introduction}

Transformers \citep{vaswani-etal-2017-transformer} have become the de-facto backbone models in several NLP applications: Machine Translation, Summarization, Question Answering, etc. By modeling recurrence relations through self-attention only, they have enabled large-scale pre-training \citep{radford-etal-2019-gpt2, brown-etal-2020-gpt3, zhang-etal-2022-opt, touvron-etal-2023-llama}. This has led to drastic advancements in Language Technologies over competing architectures, such as LSTM \citep{hochreiter-etal-1997-lstm} and GRU \citep{cho-etal-2014-gru}.

In light of the success of Transformers, several works \citep{dehghani-etal-2018-universal-transformer, yun-etal-2020-transformers-ufa, perez-etal-2021-turing-complete, merrill-sabharwal-2023-parallelism} have studied the expressivity\footnote{Expressivity of a neural network characterizes its capability in modeling complexity in data. A highly expressive network can model very complex data, while a less expressive network cannot model complex data.} of Transformers. Specifically, two lenses have been employed to study them: (\textit{a}) Lens of Universal Function Approximation, and (\textit{b}) Lens of Formal Languages and Complexity Classes. In the latter theme, several works have contributed to prove that expressivity of Transformers is upper bounded by the TC$^0$ complexity class. In the former theme, \citet{yun-etal-2020-transformers-ufa, yun-etal-2020-sparse-transformers-ufa, zaheer-etal-2020-bigbird} indicated that Transformers (and their variants) are likely to be Universal Function Approximators. However, our experimental results (Section \ref{sec:experiments-results}) \textit{contradicted} this: they are \underline{\textbf{unable}} to reliably approximate smooth\footnote{A function is smooth if it is differentiable upto some desired order, over its domain. In our case, we take functions that are differentiable upto first-order.} functions. We witnessed that Transformers could approximate the function only after a piecewise constant approximation, with \textit{large sized pieces}. Motivated by such contradicting experimental results, we ask a simple question: \textbf{Are Transformers truly Universal Function Approximators?} While we acknowledge the theoretical results by \citet{yun-etal-2020-transformers-ufa}, we find that there are further implications and analyses that provide insights into the capabilities/limitations of Transformers.

In this work, we conduct both theoretical and experimental analysis on the approximation capabilities of Transformers. We find that -- \textbf{Transformers are bad at approximating smooth functions}. We provide a relevant theoretical analysis in Section \ref{sec:expressivity-proof}, to see where the difficulty stems from, and the magnitude of this difficulty. 
To further verify our claim, we conduct experiments\footnote{Github Link: \href{https://github.com/swaroop-nath/transformer-expressivity}{\tt github.co/transformer-expressivity}} to extensively test the function approximation capabilities of the Transformer. Specifically, we conduct experiments for two axes: (\textit{a}) verifying function approximation capabilities for smooth functions, and (\textit{b}) verifying function approximation capabilities for piecewise constant functions. The results along these two axes provide a boundary of the approximation capabilities of the Transformers. We provide details on the datasets, evaluation measures, and report all our results in Section \ref{sec:experiments-results}.

Our contributions are:
\begin{enumerate}
    \item \textbf{Theoretical analysis on expressivity of Transformer Encoders }(\textbf{Theorem} \ref{thm:resolution-factor-bound}). It leads to the finding that Transformer Encoders approximate a smooth function using a piecewise constant function with very small step sizes. A smaller step size demands a large number of layers (exponential in input length) for adequate approximation, which limits the expressivity of Transformer Encoders.
    \item \textbf{Experimental analysis on expressivity of Transformers} (\textbf{Section} \ref{sec:experiments-results}). We complement our theory with experiments on the full Transformer architecture. Our experiments show that the full Transformer architecture fails at approximating smooth function, in alignment with our theory for Transformer Encoders.
\end{enumerate}

\section{Related Works}\label{sec:lit-survey}

We categorize the previous works in this line into two categories: Analyzing the expressivity of Transformers from the lens of (\textit{a})  Universal Function Approximation, and (\textit{b}) Formal Languages and Complexity Classes. We first summarize the works in these two lines, and finally provide a recap of works that question the Transformers' expressivity.

\textbf{Lens of Universal Function Approximation}.
\citet{yun-etal-2020-transformers-ufa, yun-etal-2020-sparse-transformers-ufa} provided some of the first works in analyzing Transformers as Universal Function Approximators. Notably, \citet{yun-etal-2020-transformers-ufa} contributed a step-by-step strategy for analyzing Transformers under such a lens, which has been utilized in many further works. In our work too, we build up on the strategy to show the limitations in expressivity of Transformers. \citet{zaheer-etal-2020-bigbird} used the strategy to deduce the expressivity of their new sparse-attention based Transformer. More recently, \citet{luo-etal-2022-rpe-bad} used the strategy to prove that Transformers with Relative Positional Embeddings are \underline{\textit{not}} Universal Function Approximators.

\textbf{Lens of Formal Languages}. \citet{perez-etal-2021-turing-complete, dehghani-etal-2018-universal-transformer} provided two foundational works in this line, with opposing conclusions. \citet{perez-etal-2021-turing-complete} concluded, under assumptions like arbitrary precision, that Transformers are Turing Complete. While, \citet{dehghani-etal-2018-universal-transformer} provided intuition-based arguments to claim that Transformers are not Turing Complete. \citet{bhattamishra-etal-2020-transformer-expressivity-low-lim} provide a lower limit on the expressivity of Transformers, concluding that Transformers are at least as powerful as Simplified Stateless Counter Machines (SSCM). 
On the other hand, \citet{merrill-etal-2022-saturated, hao-etal-2022-formal, merrill-sabharwal-2023-parallelism} show that the expressivity of Transformers is upper-bounded by the TC$^0$ complexity class. More recently, \citet{chiang-etal-2023-tighter-bounds} provide a tighter bound than TC$^0$ for the expressivity of Transformers.

We see that many works have probed the expressivity of Transformers from the lens of formal languages and complexity classes. Additionally, \citet{yun-etal-2020-transformers-ufa} have attempted to show that Transformers are Universal Function Approximators. However, motivated by our contradictory initial experimental results on the same, we start an investigation on the Universal Function Approximation capabilities of the Transformer. Like previous works, we provide a theoretical analysis to examine the capability. And, unlike previous works, we extensively test our claim through experiments.

\section{Notations, Definitions and Preliminaries}\label{sec:notation-definition}

\subsection{Transformer Architecture}

A Transformer \citep{vaswani-etal-2017-transformer} is a Sequence-to-Sequence mapper, composed of two main components: Encoder and Decoder. In turn, encoders and decoders are composed of several stacked blocks. Each block has two key components: \textbf{Self-Attention} and \textbf{Token-wise Feed Forward Network}, The decoder has an additional component: \textbf{Cross-Attention}. Using these blocks, Transformers essentially perform a mapping from $\mathbf{X}$ ($\in \mathbb{R}^{d \times m}$) to $\mathbf{Y}$ ($\in \mathbb{R}^{d \times n}$), where $m$ and $n$ are sequence lengths of the input and output, respectively, and $d$ is the embedding dimension.

The Self-Attention component operates according to Equation \ref{eqn:self-attn}. Specifically, it is a \textit{dot-product attention} \citep{luong-etal-2015-dot-prod-attention}, conducted across several heads (the equation depicts $h$ heads). Token-wise Feed Forward Network operates on each token according to Equation \ref{eqn:token-wise-ffn}. Cross-Attention is a variant of Self-Attention where a prefix of $\mathbf{Y}$ attends to $\mathbf{X}$, according to Equation \ref{eqn:cross-attn}.
\begin{align}
    \text{SA}(\mathbf{X}) &= \mathbf{X} + \mathbf{\mathit{W_O}} \bigoplus_{i=1}^{h}\mathbf{\mathit{W_V^i}\cdot X} \cdot \sigma\big[(\mathbf{\mathit{W_K^i}\cdot X})^T\mathbf{\mathit{W_Q^i}\cdot X}\big] \label{eqn:self-attn} \\
    \text{C}\text{A}(\mathbf{X}, \mathbf{Y}) &= \mathbf{Y^{:j}} + \mathbf{\mathit{W^{'}_O}} \bigoplus_{i=1}^{h}\mathbf{\mathit{W_V^{'i}}\cdot X} \cdot \sigma\big[(\mathbf{\mathit{W_K^{'i}}\cdot X})^T\mathbf{\mathit{W_Q^{'i}}\cdot Y^{:j}}\big] \label{eqn:cross-attn}\\
    \text{FFN}(\mathbf{X}) &= \mathbf{X} + \mathit{W}_2\cdot \text{relu}(\mathit{W}_1\cdot \mathbf{X} + \mathit{b}_1) + \mathit{b}_2\label{eqn:token-wise-ffn}
\end{align}
Where $\bigoplus_{i=1}^{h}$ denotes the concatenation operator along the embedding dimension axis, $\mathit{W^{'}_O}$, $\mathit{W_O} \> (\in \mathbb{R}^{d \times hd})$, $\mathit{W^{'}_V}$, $\mathit{W_V} \> (\in \mathbb{R}^{d \times d})$, $\mathit{W^{'}_K}$, $\mathit{W_K} \> (\in \mathbb{R}^{d \times d})$, $\mathit{W^{'}_Q}$, $\mathit{W_Q} \> (\in \mathbb{R}^{d \times d})$, $\mathit{W}_1$ ($\in \mathbb{R}^{r \times d}$), $\mathit{W}_2$ ($\in \mathbb{R}^{d \times r}$), $\mathit{b}_1$ ($\in \mathbb{R}^{r}$) and $\mathit{b}_2$ ($\in \mathbb{R}^{d}$) denote learnable matrices. $\mathbf{Y^{:j}}$ denotes the $j$-length prefix of the $n$-length sequence $\mathbf{Y}$. With these notations for the operations in a Transformer, we denote a Transformer as $\mathcal{T}^{h, d, r}$ in further sections.

\begin{figure}[h!]
    \centering
    \begin{subfigure}[b]{0.24\textwidth}
    \includegraphics[width=\linewidth]{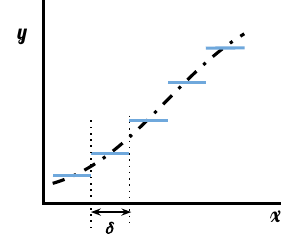}
    \caption{}
    \end{subfigure}%
    \begin{subfigure}[b]{0.24\textwidth}
    \includegraphics[width=\linewidth]{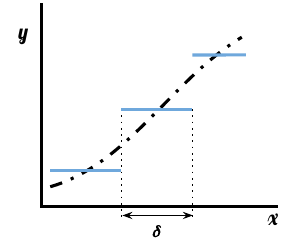}
    \caption{}
    \end{subfigure}
    \caption{Effect of changing the size of resolution factor. In (a) we have a smaller resolution factor, leading to a smaller error in approximation than (b).}
    \label{fig:effect-of-resolution-factor}
\end{figure}

\subsection{Necessary Definitions}

In this section, we provide definitions necessary for the rest of the paper. 

\begin{definition}[\textbf{Resolution Factor}]\label{def:resolution-factor}
    Let \(f\) be a function, and \(\overline{f}\) be a piecewise constant approximation to \(f\). Then, \textit{Resolution Factor}, $\delta$, is the minimum size of the pieces in \(\overline{f}\).
\end{definition}
A large $\delta$ indicates that the interval for which $\overline{f}$ is constant is large, Figure \ref{fig:effect-of-resolution-factor}. This leads to a bad approximation if $f$ has a high rate of change (derivative).
\begin{definition}[\textbf{Adequacy of Approximation}]\label{def:approximation-adequacy}
    Let \(g\) be an approximation to a function \(f\). We say that \(g\) \textit{adequately approximates} \(f\) if the following holds, for some \(\epsilon > 0\) and \(1 \leq p < \infty\).
    \[d_p(f, g) = \bigg(\int\big\|g(x) - f(x)\big\|_p^pdx\bigg)^\frac{1}{p} \leq \epsilon\]
\end{definition}
$d_p(f, g)$ defines a distance between $f$ and $g$ for different values of $x$, normalized by the $p$. Prior works \citep{yun-etal-2020-transformers-ufa, zaheer-etal-2020-bigbird} have used this definition for adequacy of approximation.
\begin{definition}[\textbf{Function with Compact Support}]\label{def:compact-support}
    A function, \(f\), is said to have a \textit{compact support} iff it is non-zero for an input from a \textit{compact set}. A set is compact iff it is \textit{closed} and \textit{bounded}.
\end{definition}

\subsection{Transformers as Universal Function Approximators}\label{subsec:ufa-transformer-steps}

In this section, we outline the $3$-step process followed by \citet{yun-etal-2020-transformers-ufa}. The work attempts to prove that Transformers, $\mathcal{T}^{h,d,r}$, approximate continuous permutation equivariant functions with compact support. We denote the set of such functions by $\mathcal{F}_{PE}$.

\textbf{Step 1}. \textbf{Approximate $\mathcal{F}_{PE}$ with piecewise constant functions}. In this step, \citet{yun-etal-2020-transformers-ufa} map the class of functions $\mathcal{F}_{PE}$ to $\mathcal{\overline{F}}_{PE}$, such that $\overline{f}$ ($\in \mathcal{\overline{F}}_{PE}$) is piecewise constant. We have seen in Definition \ref{def:resolution-factor} that the goodness of this approximation is governed by $\delta$.

\textbf{Step 2}. \textbf{Approximate $\mathcal{\overline{F}}_{PE}$ with \textit{modified} Transformers}. \citet{yun-etal-2020-transformers-ufa} show that $\mathcal{\overline{F}}_{PE}$ can be approximated by a Transformer with some simplifying modifications, such as replacing the softmax operator with the hardmax operator.

\textbf{Step 3}. \textbf{Approximate modified Transformers with (original) Transformers}. In this step, \citet{yun-etal-2020-transformers-ufa} show that the original Transformers can approximate the modified Transformers.

Using this $3$-step process, \citet{yun-etal-2020-transformers-ufa, yun-etal-2020-sparse-transformers-ufa, zaheer-etal-2020-bigbird} have analyzed the expressivity of Transformers. Additionally, \citet{yun-etal-2020-transformers-ufa} also note that with a resolution factor $\delta$, in Step 1, the number of layers in the Transformer, needed to approximate a piecewise constant function, grows as: $\mathcal{O}(m (1/\delta)^{dm})$, where $m$ is the input sequence length.

\section{Expressivity of Transformers in the Continuous Space}\label{sec:expressivity-proof}

\citet{yun-etal-2020-transformers-ufa} show that Transformers can approximate continuous functions, with a condition on the number of Transformer layers. We briefly describe their arguments in Section \ref{subsec:ufa-transformer-steps}, which show that: \textit{Transformers (}$\mathcal{T}^{h,d,r}$\textit{) approximate a piecewise constant function (}$\overline{f}$\textit{), which is, in turn, an approximation of the target function (}$f$\textit{)}. The construction of the piecewise constant approximation is governed by the resolution factor, $\delta$ (Definition \ref{def:resolution-factor}). Again, we see that this resolution factor also affects (Section \ref{subsec:ufa-transformer-steps}) the number of layers needed for an adequate approximation (Definition \ref{def:approximation-adequacy}) by the Transformer. Thus, we can deduce that: there is one factor, $\delta$, which governs the adequacy of both approximations: (\textit{a}) between $\overline{f}$ and $f$, and (\textit{b}) between $\mathcal{T}^{h,d,r}$ and $\overline{f}$. We are interested in finding how $\delta$ affects these approximations: \textbf{Does a change in $\delta$ affect both the approximations similarly, or dissimilarly}\textbf{?}\footnote{By similar effect we mean that adequacy of both approximations change in the same direction (worse or better). And, by dissimilar effect we mean that adequacy of both approximations change in the opposite directions.} We answer this in steps, by answering the following questions:

\begin{enumerate}
    \item \textbf{What governs the choice of the resolution factor, $\delta$?} We provide an intuition (Section \ref{subsec:expressivity-proof-sketch}) and a mathematical expression to answer this question (Theorem \ref{thm:resolution-factor-bound}).
    \item \textbf{How does a well-chosen resolution factor affect the conclusion of} \citet{yun-etal-2020-transformers-ufa}\textbf{, about the expressivity of Transformers?}
\end{enumerate}

\begin{theorem}\label{thm:resolution-factor-bound}
    Let \(\overline{f}\) (\(\in \mathcal{\overline{F}}\)) be an approximation to a smooth function f (\(\in \mathcal{F}\)), with goodness of approximation defined by \(d_p\)(\(\overline{f}\), \(f\)) \(\leq\) \(\epsilon\) (\(\epsilon > 0\)). Then an upper bound on \(\delta\) can be expressed as:
    \begin{equation}\label{eqn:resolution-factor-bound}
        \delta \leq \Bigg(\dfrac{2^p \cdot (p+1) \cdot \epsilon^p}{\sum_{X^0 \in \mathcal{X}^0} \sum_{i=1}^{d}\sum_{j=1}^{n} \Big|\sum_{k=1}^{m}\sum_{l=1}^{d} \frac{\partial f(X)^j_i}{\partial X_l^k}\big|_{X_0}\Big|^p}\Bigg)^{\mathlarger{\frac{1}{(p+md)}}}
    \end{equation}
where \(\mathcal{X}^0\) is a covering over the compact support (\(\mathcal{S}\)) for the function \(f\), \(p\) is the norm in \(d_p\), \(d\) is the embedding dimension, \(m\) is the input sequence length, and \(n\) is the output sequence length.
\end{theorem}

We provide proofs for Theorem \ref{thm:resolution-factor-bound} in Section \ref{sec:expressivity-proof} and Appendix \ref{asec:proof-theorem}. We include the assumptions for these proofs in Appendix \ref{asec:assumptions}. From Theorem \ref{thm:resolution-factor-bound}, we understand that for a smoothly changing function, a small resolution factor is necessary, for an adequate approximation between $\overline{f}$ and $f$. And, from Section \ref{subsec:ufa-transformer-steps}, we understand that $\mathcal{T}^{h,d,r}$ needs $\mathcal{O}(m(1/\delta)^{dm})$ layers to approximate a permutation invariant $\overline{f}$. Thus, for a smoothly changing continuous function ($f$), with a small $\delta$, large (exponential in input sequence length) number of layers are required for $\mathcal{T}^{h,d,r}$, to adequately approximate $f$. We provide a small example quantification below.

Let $\epsilon = 0.1$, $p = 1$, $d = 1$, and $|\mathcal{X}^0| = K$; for illustration, let $K \sim 10$. Assuming $f$ to be a $1$-Lipschitz smooth function in its compact support, we can have the bound $\frac{df(x)}{dx} \leq 1$. With such a setup, we have a bound on resolution factor as: $\delta \leq 0.2$. Therefore, the required number of layers for $\mathcal{T}^{h,d,r}$ grows as $\mathcal{O}(m \cdot 5^m)$. This grows very high for an input sequence length of just $10$.

From the discussion above, we understand the following:

\begin{enumerate}
    \item The choice of $\delta$ is governed by the derivative of $f$, as highlighted by Theorem \ref{thm:resolution-factor-bound}.
    \item A well-chosen $\delta$, for a smooth function, needs a large number of layers for $\mathcal{T}^{h,d,r}$.
\end{enumerate}

\textbf{What if a function is already piecewise constant?} In such a case, the resolution factor is no longer constrained by the derivative (Equation \ref{eqn:resolution-factor-bound}). Rather it depends on the size of the pieces (step-sizes). As an example, consider the following piecewise constant function, defined over the interval $[0, 1)$: $f(x) = 0.5$ for $0 \leq x < 0.75$ and $f(x) = 1$ for $0.75 \leq x < 1.0$. Here, the step-sizes are $0.75$ and $0.25$. According to Definition \ref{def:resolution-factor}, we can deduce that the resolution factor, in this case, would be $0.25$. A piecewise constant function, with small step-sizes, would still pose a challenge for Transformers. 

From the discussion, we can deduce the following things: (\textit{a}) For a smooth function, a constraint on $\delta$ is imposed as per Equation \ref{eqn:resolution-factor-bound}, and (\textit{b}) For a piecwise constant function, step-sizes constrain $\delta$, specifically $\delta = \min(step\>sizes)$.

\textbf{Limitations of our Theory}. Our theoretical treatment is based on the conclusions of \citet{yun-etal-2020-transformers-ufa}. \citet{yun-etal-2020-transformers-ufa} provide the Universal Function Approximation theories only for the Transformer Encoder, not the whole Transformer. Thus, our theory holds for the Transformer Encoder only. However, to complement our theory, and still faithfully claim our hypothesis, we perform experiments on the full Transformer. We find (Section \ref{sec:experiments-results}) that the experimental conclusions for the full Transformer align with our theoretical conclusions for the Transformer Encoder.

\subsection{Proof Sketch for Theorem \ref{thm:resolution-factor-bound}}\label{subsec:expressivity-proof-sketch}

We provide an intuition and a proof to motivate Theorem \ref{thm:resolution-factor-bound} here. From Figure \ref{fig:effect-of-resolution-factor}, we can see how varying the resolution factor can change the goodness of approximation. We see that having a smaller step size becomes necessary for functions with a large rate of change, to have a certain degree of goodness in approximation. With this intuition, we start deriving a bound on the resolution factor, based on the derivative of the function. We provide proof for the $1$-dimensional case, with $m=n=1$, using the $\ell^1$ norm. We include a proof for the general case in Appendix \ref{asec:proof-theorem}.

\begin{figure}[h]
    \centering
    \begin{subfigure}[b]{0.30\textwidth}
        \includegraphics[width=\linewidth]{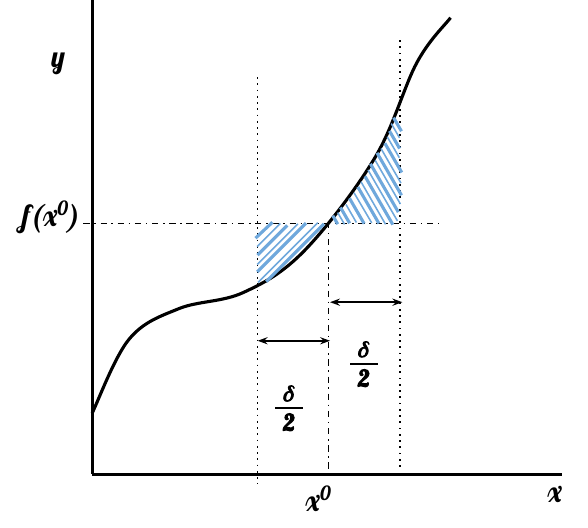}
        \caption{}
        \label{fig:continuous-vectors-transformer-proof-a}
    \end{subfigure}%
    \begin{subfigure}[b]{0.20\textwidth}
        \includegraphics[width=\linewidth]{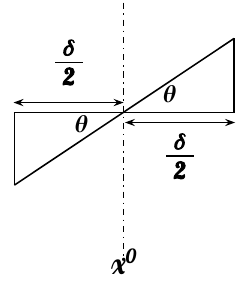}
        \caption{}
        \label{fig:continuous-vectors-transformer-proof-b}
    \end{subfigure}
    \caption{(a) Behaviour of $\overline{f}(x)$ in the $\delta/2$ neighborhood of $x^0$. In this neighborhood, we can approximate $f(x)$ by $f(x^0)$, hence $\overline{f}(x) = f(x^0)$. (b) The shaded area in $\delta/2$ neighborhood of $x^0$ for $f(x)$ can be approximated by two triangles.}
    \label{fig:continuous-vectors-transformer-proof}
\end{figure}

\begin{proof}
    We have: \(d_p\)(\(\overline{f}\), \(f\)) \(\leq \epsilon\); \(\epsilon > 0\). Qualitatively, this means that $\overline{f}$ is a good approximation of $f$.
    \[\bigg(\int_\mathcal{S}\big\|\overline{f}(x) - f(x)\big\|_p^pdx\bigg)^\frac{1}{p} \leq \epsilon\]
   Note that for $x \not\in \mathcal{S}$, $\overline{f}(x) = f(x) = 0$. With \(p = 1\), the expression becomes:
    \[\int_\mathcal{S}\big|\overline{f}(x) - f(x)\big|dx \leq \epsilon\]
    Note that the expression on the left represents the area bounded within the curves \(y=\overline{f}(x)\) and $y=f(x)$. Consider the \(\delta/2\) neighborhood of the point \(x^0\) in Figure \ref{fig:continuous-vectors-transformer-proof-a}. A good piecewise constant approximation of \(f(x)\) in this neighborhood is \(f(x^0)\), that is \(\overline{f}(x) = f(x^0)\) for \(\big(x^0 - \delta/2\big) \leq x < \big(x^0 + \delta/2\big)\). With this approximation, the expression on the left represents the area of the shaded region (Figure \ref{fig:continuous-vectors-transformer-proof-a}), for \(\big(x^0 - \delta/2\big) \leq x < \big(x^0 + \delta/2\big)\). The shaded regions can be approximated by two triangles, as shown in Figure \ref{fig:continuous-vectors-transformer-proof-b}. Thus, we have:
    \begin{align}
        \int_{x^0 - \frac{\delta}{2}}^{x^0 + \frac{\delta}{2}}\big|\overline{f}(x) - f(x)\big|dx &= 2 \cdot \dfrac{1}{2} \cdot \dfrac{\delta}{2} \cdot \bigg(\dfrac{\delta}{2}\tan \theta\bigg) + \mu_{x^0} = \dfrac{\delta^2}{4}\Bigg|\dfrac{d f(x)}{dx}\bigg|_{x^0}\Bigg| + \mu_{x^0} \nonumber
    \end{align}
    Where \(\frac{d f(x)}{dx}\Big|_{x^0}\) represents the derivative at $x^0$, and $\mu_{x^0} (\in \mathbb{R}$) is the error of approximation at $x^0$. Now, if we consider a covering \(\mathcal{X}^0\) of \(\mathcal{S}\), such that \( [x^0 - \delta/2, x^0 + \delta/2) \in \mathcal{X}^0\), we have:
    \begin{align}
        \int_\mathcal{S}\big|\overline{f}(x) - f(x)\big|dx &= \sum_{x^0 \in \mathcal{X}^0}\int_{x^0 - \frac{\delta}{2}}^{x^0 + \frac{\delta}{2}}\big|\overline{f}(x) - f(x)\big|dx = \dfrac{\delta^2}{4}\sum_{x^0 \in \mathcal{X}^0}\Bigg|\dfrac{d f(x)}{dx}\bigg|_{x^0}\Bigg| + \sum_{x^0 \in \mathcal{X}^0} \mu_{x^0} \nonumber
    \end{align}
    Thus, we have,
    \begin{align}
        \dfrac{\delta^2}{4}\sum_{x^0 \in \mathcal{X}^0}\Bigg|\dfrac{d f(x)}{dx}\bigg|_{x^0}\Bigg| + \sum_{x^0 \in \mathcal{X}^0} \mu_{x^0} \leq \epsilon \Rightarrow \delta \leq \sqrt{\dfrac{4 \epsilon - \sum_{x^0 \in \mathcal{X}^0} \mu_{x^0}}{\sum_{x^0 \in \mathcal{X}^0}\bigg|\frac{d f(x)}{dx}\Big|_{x^0}\bigg|}}\nonumber
    \end{align}
    In the expression above, $\mu_{x^0}$ is proportional to the higher order terms in the Taylor series approximation of $f(x)$ around $x^0$. Specifically, it is the fraction of area between $\overline{f}(x)$ and $f(x)$ contributed by the higher order terms. For a smooth function, if $x$ is very near around $x^0$ (that is $\delta$ is small), we can safely ignore $\mu_{x^0}$. Thus, the final expression is:
    \[\delta \leq \sqrt{\dfrac{4 \epsilon}{\sum_{x^0 \in \mathcal{X}^0}\bigg|\frac{d f(x)}{dx}\Big|_{x^0}\bigg|}}\]
    Thus, we find that if $\overline{f}$ is to be a good approximation to $f$, we have an upper bound on the resolution factor, $\delta$. The bound above matches with our expression in Equation \ref{eqn:resolution-factor-bound}. As per our intuition, we see that for functions with high rate of change, the resolution factor has to be small.
\end{proof}

\section{Experiments}\label{sec:experiments-results}
We use the Transformer Architecture proposed by \citet{vaswani-etal-2017-transformer} for our experiments. We are aware of several modifications to the vanilla architecture \citep{xiong-etal-2020-pre-ln, zaheer-etal-2020-bigbird, child-etal-2019-sparse-transformer, kitaev-etal-2020-reformer, beltagy-etal-2020-longformer}. \citet{tay-etal-2022-efficient-transformers} provide a comprehensive survey of all variants of Transformers. We refer the interested reader to that for details. These works attempt to improve stability in training and reducing computational costs, and do not focus on improving the expressivity of Transformers. Hence, it suffices to experiment on the vanilla Transformer architecture. 

We perform experiments on sythetic datasets, to ensure that data quality does not affect our conclusions. Specifically, we perform two experiments:

\begin{itemize}
    \item \textsc{Expt-I} (\textbf{Regression}): In this experiment, we train the Transformer to generate vectors to directly approximate a smooth function.
    \item \textsc{Expt-II} (\textbf{Quantized Classification}): In this experiment, we quantize the outputs to certain classes, and train the Transformer to predict the class.
\end{itemize}

\textsc{Expt-I} aims to show how Transformers fare in approximating smooth functions. In \textsc{Expt-II}, the output space is  quantized to piecewise constants, thereby essentially testing the Transformer's ability to approximate piecewise constant functions. From Section \ref{sec:expressivity-proof}, we understand that Transformers would suffer for smooth functions and small step-size piecewise constant functions. However, for piecewise constant functions with high step-sizes ($\geq 1$), Transformers can perform well. Keeping this in mind, in \textsc{Expt-II}, we train the Transformer to predict into $k$ classes, where $k$ is reasonably small. Transformer failing in \textsc{Expt-I} and succeeding in \textsc{Expt-II} would validate our hypothesis -- \textit{Transformers are bad at approximating smooth functions}. Note that, in all of our experiments, we respect the assumption of compact support (Definition \ref{def:compact-support}) for the functions.

\subsection{Evaluation Measures}

Comparing models for \textsc{Expt-I} and \textsc{Expt-II} is non-trivial, as one is a regression task and the other is a classification task. To this end, we propose a unified metric: {\tt \textbf{failure-rate}}, defined as:

{\tt \underline{failure-rate}}: The fraction of times the ground truth is not the nearest one to the generated output. 

For \textsc{Expt-II}, {\tt failure-rate} is essentially inaccuracy ($1 \> -$ accuracy). For \textsc{Expt-I}, it is computed as the fraction of times the ground truth vector is not the nearest-neighbor of the generated vector. We realize that this metric is very stringent for \textsc{Expt-I}, as there can be a lot of points in a close-by region, and also additionally propose {\tt \textbf{failure-rate@}$\mathbf{k}$}. For \textsc{Expt-II}, {\tt failure-rate@$k$} is the fraction of times the correct class is not in the top-$k$ probable outputs. For \textsc{Expt-I}, it is computed as the fraction of times the ground truth vector is not in the $k$-nearest-neighbor set of the generated vector.

The design of the metric is guided by our initial criteria of judging an approximation (Definition \ref{def:approximation-adequacy}). Mathematically, {\tt failure-rate} is linked to the Adequacy of Approximation as follows:
\begin{align}
    \text{\tt failure-rate} = \frac{1}{|\mathcal{D}|}\Big(\sum_{x \sim \mathcal{D}}\mathcal{F}(\mathcal{T}^{h,d,r}, \> x) \Big)\nonumber
\end{align}
Where $\mathcal{F}(\mathcal{T}^{h,d,r}, x) = \mathbbm{1}\big(\exists\> y \in \mathcal{D}: d_1(\mathcal{T}^{h,d,r}(x), f(y)) < d_1(\mathcal{T}^{h,d,r}(x), f(x))\big)$. $\mathbbm{1}(z)$ is $1$ if and only if $z$ evaluates to \textit{true}, and $\mathcal{D}$ is the test set.

\subsection{Synthetic Datasets \& Design Considerations}\label{subsec:synth-data-designs}

We generate a synthetic dataset using the equations in Table \ref{tab:data-gen-equations} (moved to Appendix \ref{asec:aux-datasets} for conciseness). We cover various functional forms (\textit{logarithmic}, \textit{exponential}, and \textit{polynomial}), various forms of interactions (\textit{multiplicative}, and \textit{additive}) among variables. Such coverage adequately checks the expressivity in one go, rather than designing separate datasets for various functional forms and/or interactions. We generate $200000$, $10000$ and $20000$ samples for \textit{training}, \textit{validation} and \textit{testing}, respectively. With this dataset, we ask the following empirical questions:

\begin{itemize}
    \item How does the {\tt failure-rate} vary with the number of layers in Transformer?
    \item How does the {\tt failure-rate} vary with the number of heads in the attention layer?
    \item How does the {\tt failure-rate} vary with the dimension of the Token-Wise Feed Forward Network in the Transformer?
    \item How does the {\tt failure-rate} vary with the embedding dimension of the Transformer?
    \item How does the {\tt failure-rate} vary with the number of inputs and outputs?\footnote{We include the data generation equations in \textbf{Appendix \ref{asec:aux-datasets}}.}
    \item How does the {\tt failure-rate} vary with the chosen Positional Embedding scheme?
\end{itemize}

\subsection{Implementation, Results \& Analysis}\label{subsec:experimental-results-analysis}

We use PyTorch \citep{paszke-etal-2017-pytorch} to implement Transformers. We perform hyperparameter tuning on {\tt batch size}, {\tt gradient accumulation}, {\tt epochs}, etc (see Appendix \ref{asec:training-details}). We repeat each \textit{training}-\textit{validation}-\textit{testing} pipeline $5$ times, to see the variation in performance across runs. We use \textit{mean squared error} loss for \textsc{Expt-I} and \textit{cross-entropy} loss for \textsc{Expt-II}. For \textsc{Expt-II}, we use $5$ classes for testing the Transformer across all settings. Additionally, we also report how the Transformer behaves for higher number of classes (Figure \ref{fig:trend-classes}, Appendix \ref{asec:additional-expts}).

\begin{figure*}
    \centering
    \begin{subfigure}[b]{0.46\textwidth}
            \includegraphics[width=\linewidth]{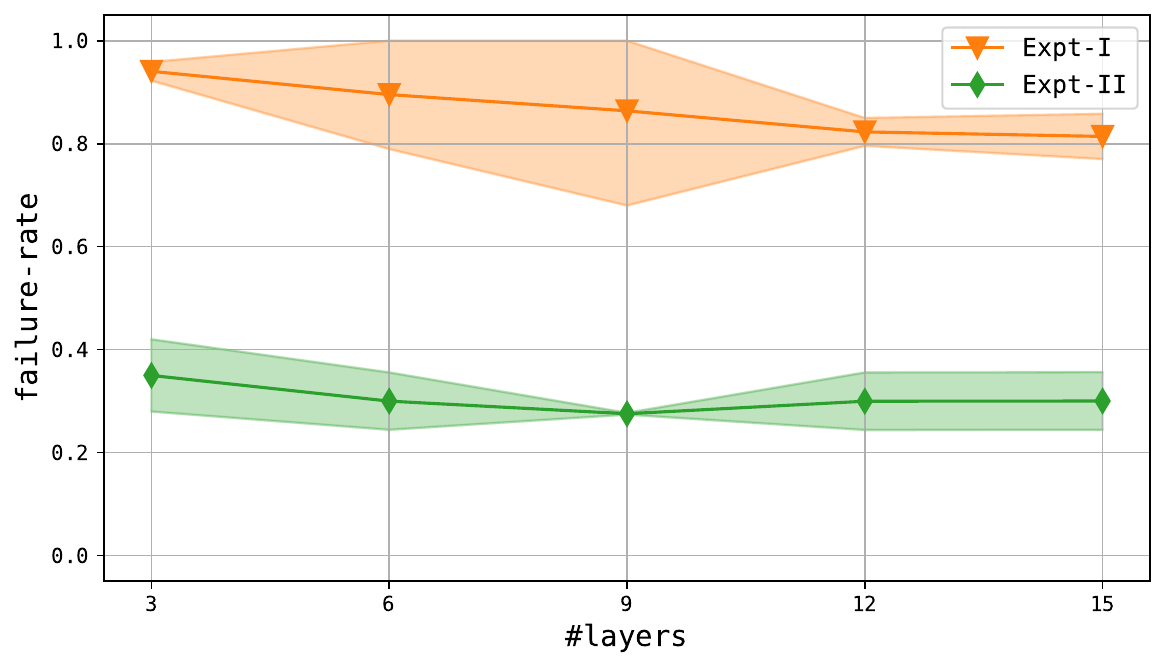}
            \caption{Trend of {\tt failure-rate} vs. $L$.}
            \label{fig:trend-layers}
    \end{subfigure}%
    \hspace{2.5em}
    \begin{subfigure}[b]{0.46\textwidth}
            \includegraphics[width=\linewidth]{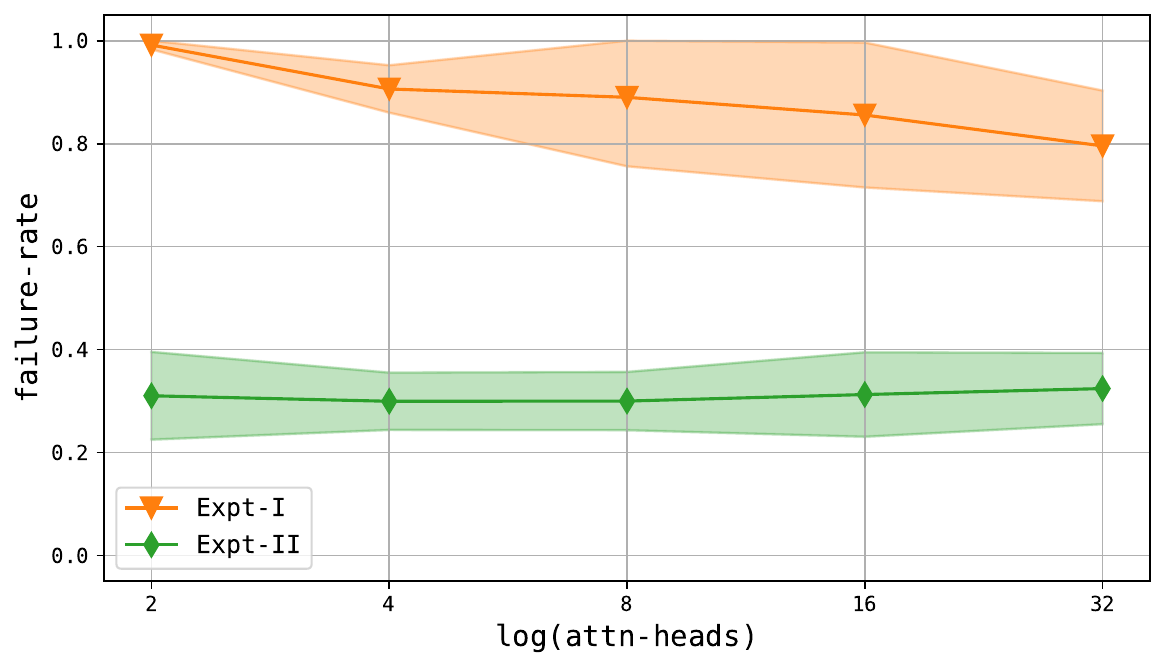}
            \caption{Trend of {\tt failure-rate} vs. $h$.}
            \label{fig:trend-attn-heads}
    \end{subfigure}
    \caption{Trend of {\tt failure-rate} with respect to the number of layers ($L$) and number of attention heads ($h$) of the Transformer. We perform each experiment (\textit{training}-\textit{validation}-\textit{testing} pipeline) $5$ times. The line in the graph corresponds to the mean across runs, and the bands around the line indicate the standard deviation. For \textsc{Expt-I}, we keep $r=d=32$, and for \textsc{Expt-II}, we keep $r=d=128$. These configurations were found to be best performing from Figure \ref{fig:trend-dims}.}\label{fig:trend-layers-attn}
\end{figure*}

\begin{figure*}
    \centering
    \begin{subfigure}[b]{0.46\textwidth}
            \includegraphics[width=\linewidth]{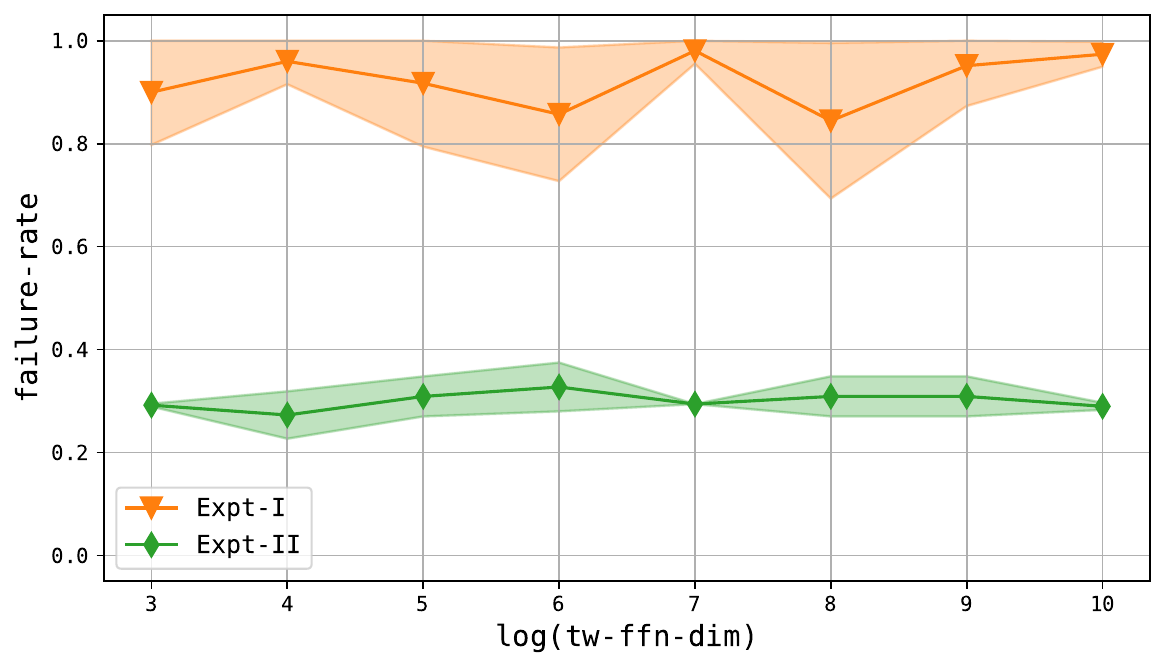}
            \caption{Trend of {\tt failure-rate} vs. $r$. We vary the latter from $8$ to $1024$, while keeping $d = 64$.}
            \label{fig:trend-dims-tw-ffn}
    \end{subfigure}%
    \hspace{2.5em}
    \begin{subfigure}[b]{0.46\textwidth}
            \includegraphics[width=\linewidth]{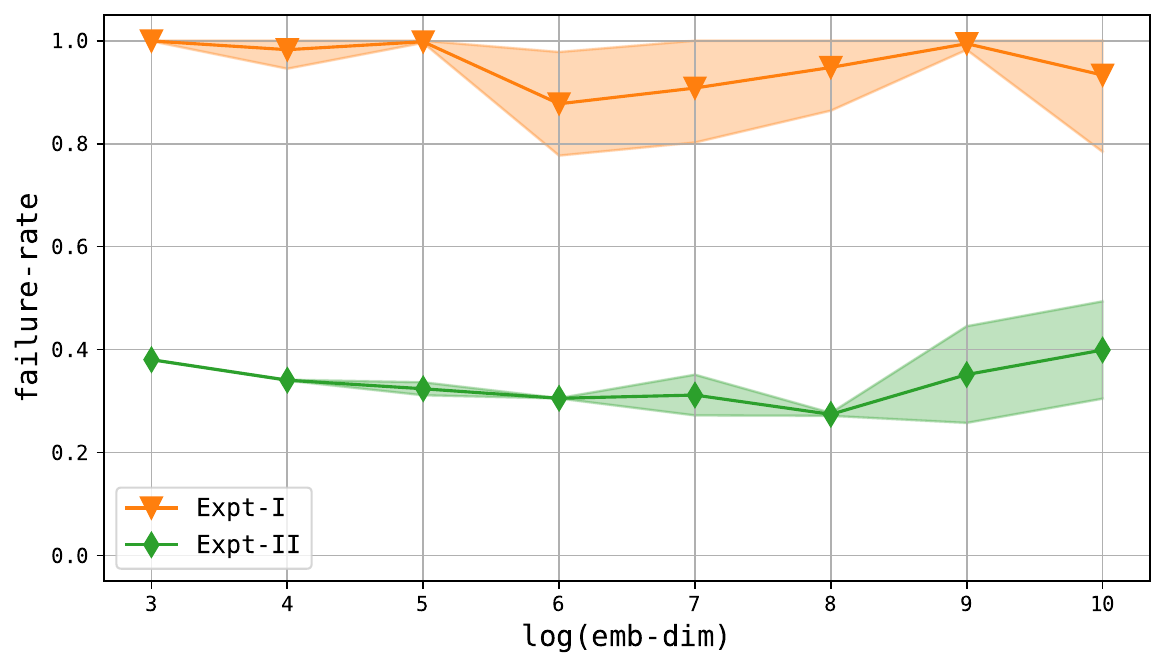}
            \caption{Trend of {\tt failure-rate} vs. $d$. We vary the latter from $4$ to $512$, while keeping $r = d$.}
            \label{fig:trend-dims-emb}
    \end{subfigure}
    \caption{Trend of {\tt failure-rate} with respect to embedding dimension ($d$) and the dimension of token-wise feed-forward network ($r$) of the Transformer. We perform each experiment (\textit{training}-\textit{validation}-\textit{testing} pipeline) $5$ times. The line in the graph corresponds to the mean across runs, and the bands around the line indicate the standard deviation.}\label{fig:trend-dims}
\end{figure*}

\textit{\textbf{Trend with Layers}}. Our theory shows that for smooth functions, a Transformer Encoder needs large number of layers. Experimentally we see that the conclusions of the theory stands for the full Transformer too. We see that for \textsc{Expt-I} (Figure \ref{fig:trend-layers}), as the number of layers increase, the {\tt failure-rate} decreases steadily. \textbf{This empirically confirms our hypothesis}. However, we also see that even with $15$ layers, the {\tt failure-rate} is very high ($\sim 0.8$) for \textsc{Expt-I}, as compared to $\sim 0.30$ for \textsc{Expt-II}. One more interesting insight is that as the number of layers increases, the standard deviation is results also decreases. 

\textit{\textbf{Trend with Attention Heads}}. In Figure \ref{fig:trend-attn-heads}, we analyze how the performance varies as the number of attention heads is changed. For \textsc{Expt-I}, we see that the performance gradually becomes better. However, the variation in performance also increases significantly. For \textsc{Expt-II}, we see that the performance and deviation remains approximately the same.

\textit{\textbf{Trend with} $\mathbf{r}$}. In Figure \ref{fig:trend-dims-tw-ffn}, we analyze the trend of performance when the token-wise feed-forward network dimension is varied. We observe that the performance for \textsc{Expt-II} does not have a discernible trend. However, for \textsc{Expt-I}, the {\tt failure-rate} oscillates a lot. Additionally, the standard deviation in performance across runs is significantly more for \textsc{Expt-I}.  In the other experiments, we make a simplifying assumption -- we keep the token-wise feed-forward dimension ($r$) equal to the embedding dimension ($d$). This helps reduce the possible space of experimental settings, and Figure \ref{fig:trend-dims-tw-ffn} shows that it is a reasonable assumption.

\textit{\textbf{Trend with} $\mathbf{d}$}. In Figure \ref{fig:trend-dims-emb}, we analyze the trend of performance against the embedding dimension of the Transformer. We see that for \textsc{Expt-II}, there is a downward trend in {\tt failure-rate} till embedding dimension $=128$. And, for \textsc{Expt-I}, there is a similar oscillatory trend. Also, as before, the standard deviation in performance remains high, thus further proving our hypothesis -- the inability of Transformers to reliably generate vectors to directly approximate a continuous function. From Figure \ref{fig:trend-dims-emb}, we understand that Transformers perform best in \textsc{Expt-I} for embedding dimension $=32$, and in \textsc{Expt-II} for embedding dimension $=128$. In the other experiments, we follow this configuration.

\begin{figure*}
    \centering
    \begin{subfigure}[b]{0.46\textwidth}
            \includegraphics[width=\linewidth]{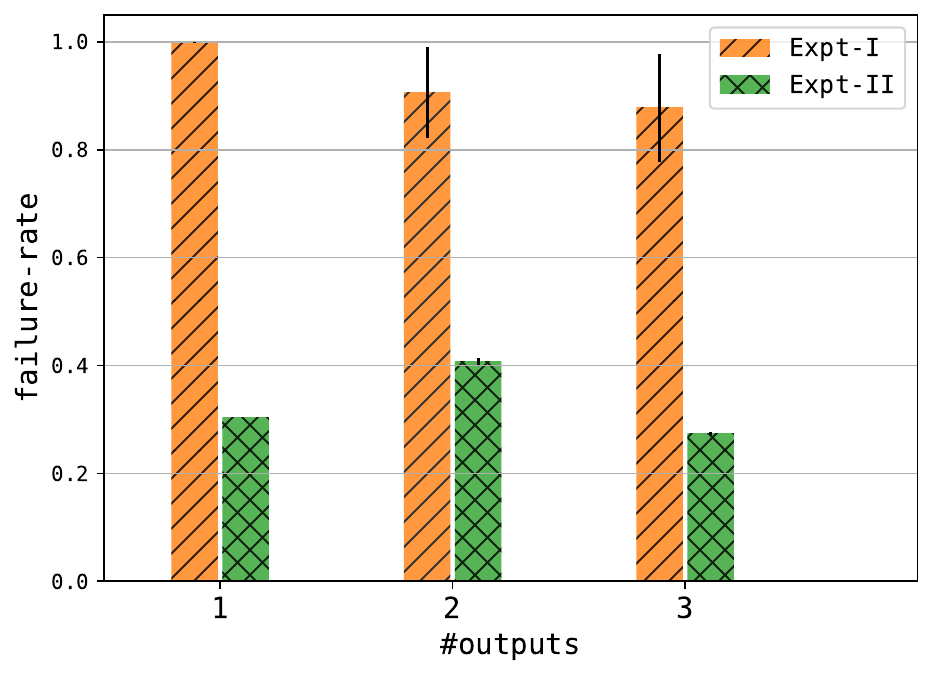}
            \caption{Trend of {\tt failure-rate} with respect to the number of outputs. We test with $1$, $2$ and $3$ outputs.}
            \label{fig:trend-outputs}
    \end{subfigure}%
    \hspace{2.5em}
    \begin{subfigure}[b]{0.46\textwidth}
            \includegraphics[width=\linewidth]{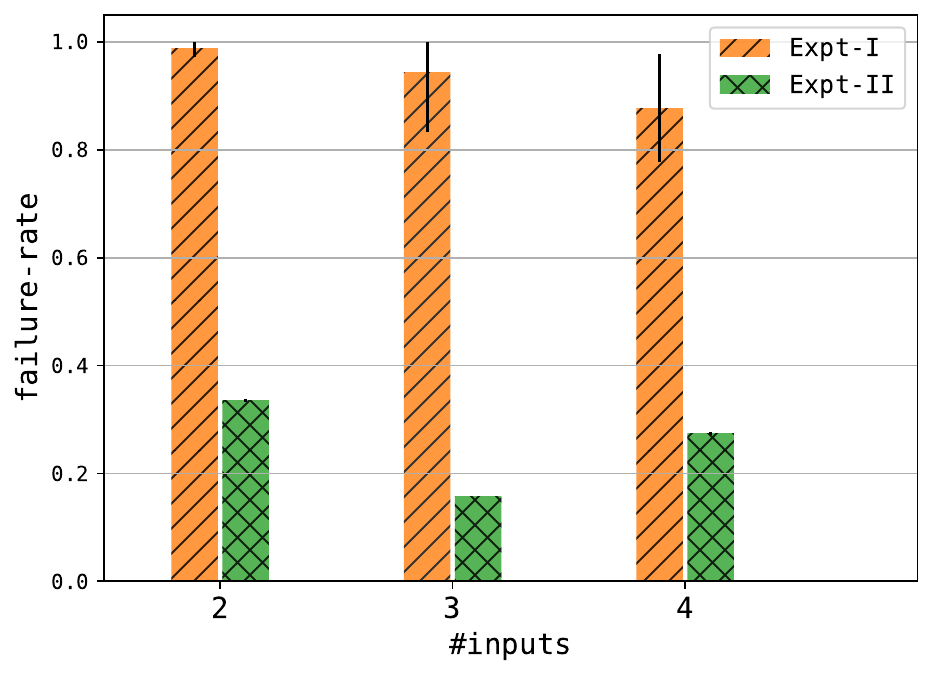}
            \caption{Trend of {\tt failure-rate} with respect to the number of inputs. We test with $2$, $3$ and $4$ inputs.}
            \label{fig:trend-inputs}
    \end{subfigure}
    \caption{Trend of {\tt failure-rate} with respect to the number of inputs and the number of outputs. We perform each experiment (\textit{training}-\textit{validation}-\textit{testing} pipeline) $5$ times. The colored bars correspond to the mean across runs, while the black lines indicate the standard deviation. For \textsc{Expt-I}, we keep $r=d=32$, and for \textsc{Expt-II}, we keep $r=d=128$. These configurations were found to be best performing from Figure \ref{fig:trend-dims}.}\label{fig:trend-inputs-outputs}
\end{figure*}

\textit{\textbf{Trend with \#inputs \& \#outputs}}. We test the Transformer by varying the number of inputs (Figure \ref{fig:trend-inputs}) and outputs (Figure \ref{fig:trend-outputs}). We see that as the \#inputs or \#outputs is increased, the performance for \textsc{Expt-I} becomes better. However, the variation in the performance also increases significantly. Additionally, we also see that irrespective of \#inputs and \#outputs, the huge performance gap between \textsc{Expt-I} and \textsc{Expt-II} still remains.

\begin{figure}[h!]
    \centering
    \includegraphics[width=0.5\textwidth]{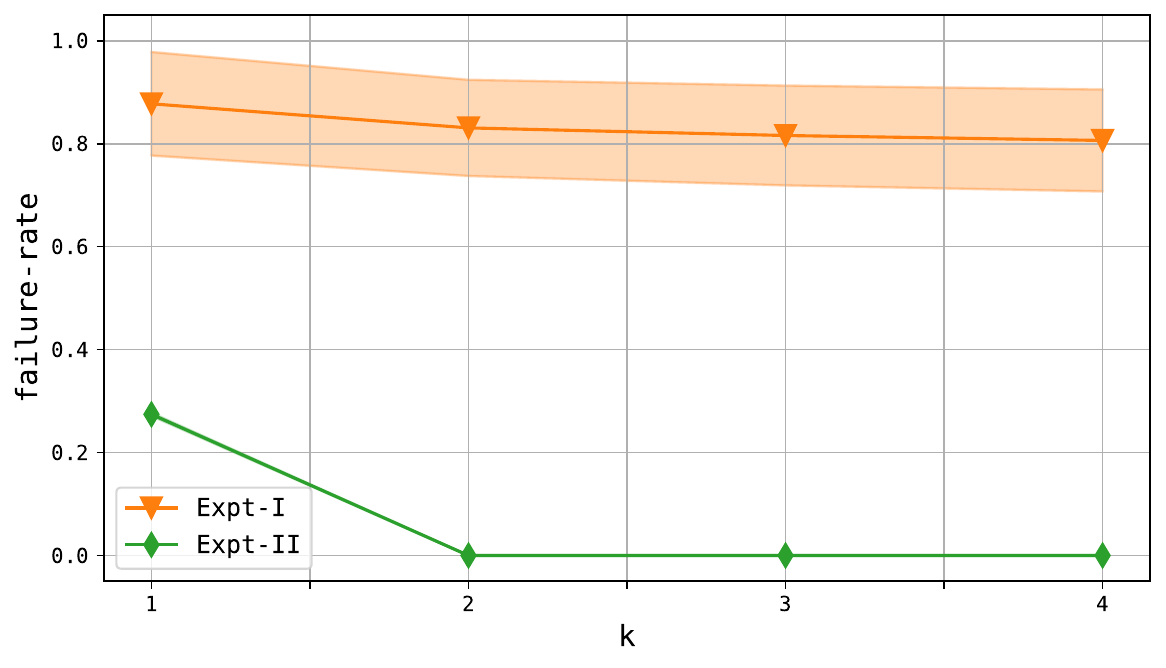}
    \caption{Trend of performance of Transformer for \textsc{Expt-I} and \textsc{Expt-II} for various $k$ in {\tt failure-rate@}$k$. For \textsc{Expt-I}, we keep $r=d=32$, and for \textsc{Expt-II}, we keep $r=d=128$. These configurations were found to be best performing from Figure \ref{fig:trend-dims}.}
    \label{fig:trend-frate-vs-k}
\end{figure}

We had stated earlier, {\tt failure-rate} can be a stringent metric to judge performance in \textsc{Expt-II}. To alleviate the situation, we proposed {\tt failure-rate@}$k$. We examine how the performance varies across various values of $k$ for \textsc{Expt-I} and \textsc{Expt-II}, Figure \ref{fig:trend-frate-vs-k}. We see that even if $k$ is increased, the Transformer fails significantly for \textsc{Expt-I}.

We report the trend wrt positional embedding and the effect of data size on performance in Appendix \ref{asec:additional-expts}. We also include \textsc{t-sne} plots (Figure \ref{fig:t-sne-y*}) for the generated and ground truth points, to show how the Transformer fails for smooth functions.

\section{Summary, Conclusion and Future Work}\label{sec:sum-conc}
In this work, we analyze the function approximation capabilities of the Transformer. We provide a theoretical treatment, analyzing the capability, in Section \ref{sec:expressivity-proof}. We prove that Transformer Encoders cannot practically approximate smooth functions. Following that, we provide experimental results on several settings to evaluate the capability of the full Transformer, in Section \ref{sec:experiments-results}. Experimentally, we complement our theoretical findings, and show that the full Transformer also cannot practically approximate smooth functions. Based on our theoretical and experimental findings, we thus conclude that: \textbf{Transformers are bad at approximating smooth functions}. However, \textit{they are quite adept at approximating piecewise constant functions with moderately large-sized pieces}. Given this finding, our future work would attempt to: (\textit{a}) Pinpoint the source of this limited expressivity, and (\textit{b}) Modify the Transformer architecture to attain Universal Function Approximation. Additionally, for interested readers, we provide a few more future works along with limitations of our work in Appendix \ref{asec:limitations}.

\bibliography{references}

\begin{thebibliography}{26}
\providecommand{\natexlab}[1]{#1}
\providecommand{\url}[1]{\texttt{#1}}
\expandafter\ifx\csname urlstyle\endcsname\relax
  \providecommand{\doi}[1]{doi: #1}\else
  \providecommand{\doi}{doi: \begingroup \urlstyle{rm}\Url}\fi

\bibitem[Beltagy et~al.(2020)Beltagy, Peters, and Cohan]{beltagy-etal-2020-longformer}
Iz~Beltagy, Matthew~E. Peters, and Arman Cohan.
\newblock Longformer: The long-document transformer.
\newblock \emph{CoRR}, abs/2004.05150, 2020.
\newblock URL \url{https://arxiv.org/abs/2004.05150}.

\bibitem[Bhattamishra et~al.(2020)Bhattamishra, Ahuja, and Goyal]{bhattamishra-etal-2020-transformer-expressivity-low-lim}
Satwik Bhattamishra, Kabir Ahuja, and Navin Goyal.
\newblock On the {A}bility and {L}imitations of {T}ransformers to {R}ecognize {F}ormal {L}anguages.
\newblock In Bonnie Webber, Trevor Cohn, Yulan He, and Yang Liu, editors, \emph{Proceedings of the 2020 Conference on Empirical Methods in Natural Language Processing (EMNLP)}, pages 7096--7116, Online, November 2020. Association for Computational Linguistics.
\newblock \doi{10.18653/v1/2020.emnlp-main.576}.
\newblock URL \url{https://aclanthology.org/2020.emnlp-main.576}.

\bibitem[Brown et~al.(2020)Brown, Mann, Ryder, Subbiah, Kaplan, Dhariwal, Neelakantan, Shyam, Sastry, Askell, Agarwal, Herbert{-}Voss, Krueger, Henighan, Child, Ramesh, Ziegler, Wu, Winter, Hesse, Chen, Sigler, Litwin, Gray, Chess, Clark, Berner, McCandlish, Radford, Sutskever, and Amodei]{brown-etal-2020-gpt3}
Tom~B. Brown, Benjamin Mann, Nick Ryder, Melanie Subbiah, Jared Kaplan, Prafulla Dhariwal, Arvind Neelakantan, Pranav Shyam, Girish Sastry, Amanda Askell, Sandhini Agarwal, Ariel Herbert{-}Voss, Gretchen Krueger, Tom Henighan, Rewon Child, Aditya Ramesh, Daniel~M. Ziegler, Jeffrey Wu, Clemens Winter, Christopher Hesse, Mark Chen, Eric Sigler, Mateusz Litwin, Scott Gray, Benjamin Chess, Jack Clark, Christopher Berner, Sam McCandlish, Alec Radford, Ilya Sutskever, and Dario Amodei.
\newblock Language models are few-shot learners.
\newblock \emph{CoRR}, abs/2005.14165, 2020.
\newblock URL \url{https://arxiv.org/abs/2005.14165}.

\bibitem[Chiang et~al.(2023)Chiang, Cholak, and Pillay]{chiang-etal-2023-tighter-bounds}
David Chiang, Peter Cholak, and Anand Pillay.
\newblock Tighter bounds on the expressivity of transformer encoders.
\newblock In \emph{Proc. ICML}, pages 5544--5562, 2023.

\bibitem[Child et~al.(2019)Child, Gray, Radford, and Sutskever]{child-etal-2019-sparse-transformer}
Rewon Child, Scott Gray, Alec Radford, and Ilya Sutskever.
\newblock Generating long sequences with sparse transformers.
\newblock \emph{CoRR}, abs/1904.10509, 2019.
\newblock URL \url{http://arxiv.org/abs/1904.10509}.

\bibitem[Cho et~al.(2014)Cho, van Merri{\"e}nboer, Bahdanau, and Bengio]{cho-etal-2014-gru}
Kyunghyun Cho, Bart van Merri{\"e}nboer, Dzmitry Bahdanau, and Yoshua Bengio.
\newblock On the properties of neural machine translation: Encoder{--}decoder approaches.
\newblock In Dekai Wu, Marine Carpuat, Xavier Carreras, and Eva~Maria Vecchi, editors, \emph{Proceedings of {SSST}-8, Eighth Workshop on Syntax, Semantics and Structure in Statistical Translation}, pages 103--111, Doha, Qatar, October 2014. Association for Computational Linguistics.
\newblock \doi{10.3115/v1/W14-4012}.
\newblock URL \url{https://aclanthology.org/W14-4012}.

\bibitem[Dehghani et~al.(2018)Dehghani, Gouws, Vinyals, Uszkoreit, and Kaiser]{dehghani-etal-2018-universal-transformer}
Mostafa Dehghani, Stephan Gouws, Oriol Vinyals, Jakob Uszkoreit, and Lukasz Kaiser.
\newblock Universal transformers.
\newblock \emph{ArXiv}, abs/1807.03819, 2018.
\newblock URL \url{https://api.semanticscholar.org/CorpusID:49667762}.

\bibitem[Devlin et~al.(2019)Devlin, Chang, Lee, and Toutanova]{devlin-etal-2019-bert}
Jacob Devlin, Ming-Wei Chang, Kenton Lee, and Kristina Toutanova.
\newblock {BERT}: Pre-training of deep bidirectional transformers for language understanding.
\newblock In Jill Burstein, Christy Doran, and Thamar Solorio, editors, \emph{Proceedings of the 2019 Conference of the North {A}merican Chapter of the Association for Computational Linguistics: Human Language Technologies, Volume 1 (Long and Short Papers)}, pages 4171--4186, Minneapolis, Minnesota, June 2019. Association for Computational Linguistics.
\newblock \doi{10.18653/v1/N19-1423}.
\newblock URL \url{https://aclanthology.org/N19-1423}.

\bibitem[Hao et~al.(2022)Hao, Angluin, and Frank]{hao-etal-2022-formal}
Yiding Hao, Dana Angluin, and Robert Frank.
\newblock Formal language recognition by hard attention transformers: Perspectives from circuit complexity.
\newblock \emph{Transactions of the Association for Computational Linguistics}, 10:\penalty0 800--810, 2022.
\newblock \doi{10.1162/tacl_a_00490}.
\newblock URL \url{https://aclanthology.org/2022.tacl-1.46}.

\bibitem[Hochreiter and Schmidhuber(1997)]{hochreiter-etal-1997-lstm}
Sepp Hochreiter and J{\"u}rgen Schmidhuber.
\newblock Long short-term memory.
\newblock \emph{Neural computation}, 9\penalty0 (8):\penalty0 1735--1780, 1997.

\bibitem[Kitaev et~al.(2020)Kitaev, Kaiser, and Levskaya]{kitaev-etal-2020-reformer}
Nikita Kitaev, Lukasz Kaiser, and Anselm Levskaya.
\newblock Reformer: The efficient transformer.
\newblock \emph{CoRR}, abs/2001.04451, 2020.
\newblock URL \url{https://arxiv.org/abs/2001.04451}.

\bibitem[Luo et~al.(2022)Luo, Li, Zheng, Liu, Wang, and He]{luo-etal-2022-rpe-bad}
Shengjie Luo, Shanda Li, Shuxin Zheng, Tie-Yan Liu, Liwei Wang, and Di~He.
\newblock Your transformer may not be as powerful as you expect.
\newblock In S.~Koyejo, S.~Mohamed, A.~Agarwal, D.~Belgrave, K.~Cho, and A.~Oh, editors, \emph{Advances in Neural Information Processing Systems}, volume~35, pages 4301--4315. Curran Associates, Inc., 2022.
\newblock URL \url{https://proceedings.neurips.cc/paper_files/paper/2022/file/1ba5f64159d67775a251cf9ce386a2b9-Paper-Conference.pdf}.

\bibitem[Luong et~al.(2015)Luong, Pham, and Manning]{luong-etal-2015-dot-prod-attention}
Thang Luong, Hieu Pham, and Christopher~D. Manning.
\newblock Effective approaches to attention-based neural machine translation.
\newblock In Llu{\'\i}s M{\`a}rquez, Chris Callison-Burch, and Jian Su, editors, \emph{Proceedings of the 2015 Conference on Empirical Methods in Natural Language Processing}, pages 1412--1421, Lisbon, Portugal, September 2015. Association for Computational Linguistics.
\newblock \doi{10.18653/v1/D15-1166}.
\newblock URL \url{https://aclanthology.org/D15-1166}.

\bibitem[Merrill and Sabharwal(2023)]{merrill-sabharwal-2023-parallelism}
William Merrill and Ashish Sabharwal.
\newblock The parallelism tradeoff: Limitations of log-precision transformers.
\newblock \emph{Transactions of the Association for Computational Linguistics}, 11:\penalty0 531--545, 2023.
\newblock \doi{10.1162/tacl_a_00562}.
\newblock URL \url{https://aclanthology.org/2023.tacl-1.31}.

\bibitem[Merrill et~al.(2022)Merrill, Sabharwal, and Smith]{merrill-etal-2022-saturated}
William Merrill, Ashish Sabharwal, and Noah~A. Smith.
\newblock Saturated transformers are constant-depth threshold circuits.
\newblock \emph{Transactions of the Association for Computational Linguistics}, 10:\penalty0 843--856, 2022.
\newblock \doi{10.1162/tacl_a_00493}.
\newblock URL \url{https://aclanthology.org/2022.tacl-1.49}.

\bibitem[Paszke et~al.(2017)Paszke, Gross, Chintala, Chanan, Yang, DeVito, Lin, Desmaison, Antiga, and Lerer]{paszke-etal-2017-pytorch}
Adam Paszke, Sam Gross, Soumith Chintala, Gregory Chanan, Edward Yang, Zachary DeVito, Zeming Lin, Alban Desmaison, Luca Antiga, and Adam Lerer.
\newblock Automatic differentiation in pytorch.
\newblock In \emph{NIPS-W}, 2017.

\bibitem[Perez et~al.(2021)Perez, BarcelÃ³, and Marinkovic]{perez-etal-2021-turing-complete}
Jorge Perez, Pablo BarcelÃ³, and Javier Marinkovic.
\newblock Attention is turing-complete.
\newblock \emph{Journal of Machine Learning Research}, 22\penalty0 (75):\penalty0 1--35, 2021.
\newblock URL \url{http://jmlr.org/papers/v22/20-302.html}.

\bibitem[Radford et~al.(2019)Radford, Wu, Child, Luan, Amodei, and Sutskever]{radford-etal-2019-gpt2}
Alec Radford, Jeff Wu, Rewon Child, David Luan, Dario Amodei, and Ilya Sutskever.
\newblock Language models are unsupervised multitask learners.
\newblock 2019.
\newblock URL \url{https://api.semanticscholar.org/CorpusID:160025533}.

\bibitem[Tay et~al.(2022)Tay, Dehghani, Bahri, and Metzler]{tay-etal-2022-efficient-transformers}
Yi~Tay, Mostafa Dehghani, Dara Bahri, and Donald Metzler.
\newblock Efficient transformers: A survey.
\newblock \emph{ACM Comput. Surv.}, 55\penalty0 (6), dec 2022.
\newblock ISSN 0360-0300.
\newblock \doi{10.1145/3530811}.
\newblock URL \url{https://doi.org/10.1145/3530811}.

\bibitem[Touvron et~al.(2023)Touvron, Lavril, Izacard, Martinet, Lachaux, Lacroix, Rozière, Goyal, Hambro, Azhar, Rodriguez, Joulin, Grave, and Lample]{touvron-etal-2023-llama}
Hugo Touvron, Thibaut Lavril, Gautier Izacard, Xavier Martinet, Marie-Anne Lachaux, Timothée Lacroix, Baptiste Rozière, Naman Goyal, Eric Hambro, Faisal Azhar, Aurelien Rodriguez, Armand Joulin, Edouard Grave, and Guillaume Lample.
\newblock Llama: Open and efficient foundation language models, 2023.

\bibitem[Vaswani et~al.(2017)Vaswani, Shazeer, Parmar, Uszkoreit, Jones, Gomez, Kaiser, and Polosukhin]{vaswani-etal-2017-transformer}
Ashish Vaswani, Noam Shazeer, Niki Parmar, Jakob Uszkoreit, Llion Jones, Aidan~N Gomez, \L~ukasz Kaiser, and Illia Polosukhin.
\newblock Attention is all you need.
\newblock In I.~Guyon, U.~Von Luxburg, S.~Bengio, H.~Wallach, R.~Fergus, S.~Vishwanathan, and R.~Garnett, editors, \emph{Advances in Neural Information Processing Systems}, volume~30. Curran Associates, Inc., 2017.
\newblock URL \url{https://proceedings.neurips.cc/paper_files/paper/2017/file/3f5ee243547dee91fbd053c1c4a845aa-Paper.pdf}.

\bibitem[Xiong et~al.(2020)Xiong, Yang, He, Zheng, Zheng, Xing, Zhang, Lan, Wang, and Liu]{xiong-etal-2020-pre-ln}
Ruibin Xiong, Yunchang Yang, Di~He, Kai Zheng, Shuxin Zheng, Chen Xing, Huishuai Zhang, Yanyan Lan, Liwei Wang, and Tieyan Liu.
\newblock On layer normalization in the transformer architecture.
\newblock In Hal~Daumé III and Aarti Singh, editors, \emph{Proceedings of the 37th International Conference on Machine Learning}, volume 119 of \emph{Proceedings of Machine Learning Research}, pages 10524--10533. PMLR, 13--18 Jul 2020.
\newblock URL \url{https://proceedings.mlr.press/v119/xiong20b.html}.

\bibitem[Yun et~al.(2020{\natexlab{a}})Yun, Bhojanapalli, Rawat, Reddi, and Kumar]{yun-etal-2020-transformers-ufa}
Chulhee Yun, Srinadh Bhojanapalli, Ankit~Singh Rawat, Sashank Reddi, and Sanjiv Kumar.
\newblock Are transformers universal approximators of sequence-to-sequence functions?
\newblock In \emph{International Conference on Learning Representations}, 2020{\natexlab{a}}.

\bibitem[Yun et~al.(2020{\natexlab{b}})Yun, Chang, Bhojanapalli, Rawat, Reddi, and Kumar]{yun-etal-2020-sparse-transformers-ufa}
Chulhee Yun, Yin-Wen Chang, Srinadh Bhojanapalli, Ankit~Singh Rawat, Sashank Reddi, and Sanjiv Kumar.
\newblock O(n) connections are expressive enough: Universal approximability of sparse transformers.
\newblock In H.~Larochelle, M.~Ranzato, R.~Hadsell, M.F. Balcan, and H.~Lin, editors, \emph{Advances in Neural Information Processing Systems}, volume~33, pages 13783--13794. Curran Associates, Inc., 2020{\natexlab{b}}.
\newblock URL \url{https://proceedings.neurips.cc/paper_files/paper/2020/file/9ed27554c893b5bad850a422c3538c15-Paper.pdf}.

\bibitem[Zaheer et~al.(2020)Zaheer, Guruganesh, Dubey, Ainslie, Alberti, Ontanon, Pham, Ravula, Wang, Yang, and Ahmed]{zaheer-etal-2020-bigbird}
Manzil Zaheer, Guru Guruganesh, Kumar~Avinava Dubey, Joshua Ainslie, Chris Alberti, Santiago Ontanon, Philip Pham, Anirudh Ravula, Qifan Wang, Li~Yang, and Amr Ahmed.
\newblock Big bird: Transformers for longer sequences.
\newblock In H.~Larochelle, M.~Ranzato, R.~Hadsell, M.F. Balcan, and H.~Lin, editors, \emph{Advances in Neural Information Processing Systems}, volume~33, pages 17283--17297. Curran Associates, Inc., 2020.
\newblock URL \url{https://proceedings.neurips.cc/paper_files/paper/2020/file/c8512d142a2d849725f31a9a7a361ab9-Paper.pdf}.

\bibitem[Zhang et~al.(2022)Zhang, Roller, Goyal, Artetxe, Chen, Chen, Dewan, Diab, Li, Lin, Mihaylov, Ott, Shleifer, Shuster, Simig, Koura, Sridhar, Wang, and Zettlemoyer]{zhang-etal-2022-opt}
Susan Zhang, Stephen Roller, Naman Goyal, Mikel Artetxe, Moya Chen, Shuohui Chen, Christopher Dewan, Mona Diab, Xian Li, Xi~Victoria Lin, Todor Mihaylov, Myle Ott, Sam Shleifer, Kurt Shuster, Daniel Simig, Punit~Singh Koura, Anjali Sridhar, Tianlu Wang, and Luke Zettlemoyer.
\newblock Opt: Open pre-trained transformer language models, 2022.

\end{thebibliography}
\bibliographystyle{plainnat}

\appendix

\section{Limitations}\label{asec:limitations}

In this work, we extensively probe the function approximation capabilities of Transformers. First we theoretically show that Transformer Encoders are bad at approximating smooth functions. Then we experimentally show that the full Transformer architecture is also bad at approximating smooth functions, complementing our theory. However, our work has the following limitations:

\begin{enumerate}
    \item We do not theoretically analyse the function approximation capabilities of the full Transformer architecture. However, we believe that the presented analysis (theory $+$ experiments) sufficiently supports our hypothesis.
    \item We highlight the limited expressivity of Transformers for smooth functions. However, we do not dive deeper into highlighting the varying extent of difficulty faced by the Transformer for different functions. For instance, a Transformer might be better at approximating \textit{logarithmic functions} than \textit{exponential functions}. We leave this as a future work.
    \item We do not provide any modification to the Transformer architecture to improve its expressivity. We leave this as a future work.
\end{enumerate}

\section{Assumptions of our Proofs}\label{asec:assumptions}

Our proofs, presented in Section \ref{subsec:expressivity-proof-sketch} and Appendix \ref{asec:proof-theorem}, are based on the following assumptions:

\begin{enumerate}
    \item \textit{\textbf{Assumption}} $\mathbf{\mathit{1}}$: \textit{The function to be approximated is smooth, which has continuous first-order derivatives over the entire compact support (}$\mathcal{S}$\textit{) of the function}. 
    \item[] \textbf{\textit{Effect}}: This assumption limits the exploration space in deciding which functions cannot be reliably approximated by Transformers to smooth functions. We admit that the boundary may be broader, that is there can be non-smooth continuous functions, which are difficult for Transformers to approximate.
    \item \textit{\textbf{Assumption}} $\mathbf{\mathit{2}}$: \textit{The higher order terms in the Taylor series approximation (}$\mu$\textit{) are negligible.}  
    \item[] \textbf{\textit{Validity}}: This assumption is quite valid for smooth functions when the approximation region is small ($X$ is near the approximation point $X^0$). In our case, $X$ is indeed near $X^0$ (owing to small $\delta$).
\end{enumerate}

\section{Proof for Theorem \ref{thm:resolution-factor-bound}}\label{asec:proof-theorem}

In this section, we prove \ref{thm:resolution-factor-bound} for the general case, where  $f(X) \in \mathbb{R}^{d \times n}$, $n$ is the number of output variables and $X \in \mathbb{R}^{d \times m}$, $m$ is the number of input variables.
\begin{proof}
    We have: \(d_p(\overline{f}, f) \leq \epsilon\). Qualitatively, this means that $\overline{f}$ is a good approximation of $f$.
    \[\bigg(\int_\mathcal{S}\big\|\overline{f}(X) - f(X)\big\|_p^pdx\bigg)^\mathlarger{\frac{1}{p}} \leq \epsilon\]
   Note that for $X \not\in \mathcal{S}$, $\overline{f}(X) = f(X) = 0$. Now, using element-wise norm of a matrix,
    \begin{align}
        \big\|\overline{f}(X) - f(X)\big\|_p^p &= \bigg\{\sum_{i=1}^{d}\sum_{j=1}^{n}\big|\overline{f}(X)_i^j - f(X)_i^j\big|^p\bigg\}^{p \times \mathlarger{\frac{1}{p}}} = \sum_{i=1}^{d}\sum_{j=1}^{n}\big|\overline{f}(X)_i^j - f(X)_i^j\big|^p  \nonumber
    \end{align}
    So, we have,
    \begin{align}
        \bigg(\int_\mathcal{S}\big\|\overline{f}(X) - f(X)\big\|_p^pdx\bigg)^\mathlarger{\frac{1}{p}} &\leq \epsilon \nonumber \\
        \Rightarrow \bigg( \int_\mathcal{S} \sum_{i=1}^{d}\sum_{j=1}^{n}\big|\overline{f}(X)_i^j - f(X)_i^j\big|^p dX \bigg)^\mathlarger{\frac{1}{p}} &\leq \epsilon \nonumber
    \end{align}    
    We look at the $\delta/2$ neighborhood of $X^0$. Note that the neighborhood is a hypercube. As before, we have $\overline{f}(X)^j = f(X^0)^j$ in this neighborhood. Again, in this neighborhood, $f(X)^j$ can be rewritten as:
    \[f(X)_i^j = f(X^0)_i^j + \sum_{k=1}^{m}\sum_{l=1}^{d} \dfrac{\partial f(X)^j_i}{\partial X_l^k}\bigg|_{X_0} \cdot (X_l^k - X_l^{k, 0}) + \mu_{X^0}^{i,j} \quad \forall i, j\]
    Where $X_l^k$ denotes the $l$-th component of the $k$-th sequence item in $X$, and $\mu_{X^0}^{i,j}$ denotes the error of approximation for $f(X)_i^j$ around $X^0$. Note that this approximation is the linear approximation of a function using first-order partial derivatives. For a function $z = g(x, y)$, we have the approximation as: $g(x_0 + \Delta x, y_0 + \Delta y) \approx g(x_0, y_0) + \frac{\partial g}{\partial x} \cdot \Delta x + \frac{\partial g}{\partial y} \cdot \Delta y$. In our case, $X \in \mathbb{R}^{d \times m}$, hence we have two summations: one over $d$ and another over $m$. With $\overline{f}(X)^j = f(X^0)^j$, we have,
    \[\overline{f}(X)_i^j - f(X)_i^j = -\sum_{k=1}^{m}\sum_{l=1}^{d} \dfrac{\partial f(X)^j_i}{\partial X_l^k}\bigg|_{X_0} \cdot (X_l^k - X_l^{k, 0}) - \mu_{X^0}^{i,j} \quad \forall i, j\]
    In the following mathematical statements, we will use $A$ to denote $\sum_{k=1}^{m}\sum_{l=1}^{d} \dfrac{\partial f(X)^j_i}{\partial X_l^k}\bigg|_{X_0} \cdot (X_l^k - X_l^{k, 0})$, for conciseness. We will perform appropriate substitution as and when necessary. Thus, in the $\delta/2$ neighborhood of $X^0$, we can compute the $\ell^p$ error of approximation as ($[\delta/2]_{m,d}$ denotes a $d \times m$-dimensional matrix, with each element as $\delta/2$):
    \begin{align}
        &\int_{X^0 - [\delta/2]_{m,d}}^{X^0 + [\delta/2]_{m,d}} \sum_{i=1}^{d}\sum_{j=1}^{n}\big|\overline{f}(X)_i^j - f(X)_i^j\big|^p dX \nonumber \\\nonumber\\
        = &\int_{X^0 - [\delta/2]_{m,d}}^{X^0 + [\delta/2]_{m,d}} \sum_{i=1}^{d}\sum_{j=1}^{n}\Bigg|A + \mu_{X^0}^{i,j}\Bigg|^p dX \nonumber \\\nonumber\\
        = &\int_{X^0 - [\delta/2]_{m,d}}^{X^0 + [\delta/2]_{m,d}} \sum_{i=1}^{d}\sum_{j=1}^{n} \Bigg|A \cdot \bigg(1 + \dfrac{\mu_{X^0}^{i,j}}{A}\bigg) \Bigg|^p dX \nonumber
    \end{align}
    $\mu_{X^0}^{i, j}$ represents the higher order terms in Taylor series expansion of $f(X)^j_i$ around $X^0$. Again, as in Section \ref{sec:expressivity-proof}, we can safely ignore this. Thus, we have $1 + \dfrac{\mu_{X^0}^{i,j}}{A} \approx 1$. Now, we have:
    \begin{align}
        &\int_{X^0 - [\delta/2]_{m,d}}^{X^0 + [\delta/2]_{m,d}}\sum_{i=1}^{d}\sum_{j=1}^{n} |A|^p dX \nonumber \\\nonumber \\
        = &\int_{X^0 - [\delta/2]_{m,d}}^{X^0 + [\delta/2]_{m,d}}\sum_{i=1}^{d}\sum_{j=1}^{n}\Bigg|\sum_{k=1}^{m}\sum_{l=1}^{d} \dfrac{\partial f(X)^j_i}{\partial X_l^k}\bigg|_{X_0} \cdot (X_l^k - X_l^{k, 0}) \Bigg|^p dX \nonumber \\\nonumber \\
        =&\sum_{i=1}^{d}\sum_{j=1}^{n}\int_{X^0 - [\delta/2]_{m,d}}^{X^0 + [\delta/2]_{m,d}}\Bigg|\sum_{k=1}^{m}\sum_{l=1}^{d} \dfrac{\partial f(X)^j_i}{\partial X_l^k}\bigg|_{X_0} \cdot (X_l^k - X_l^{k, 0}) \Bigg|^p dX \nonumber
    \end{align}
    The integral can be computed as follows (we refer to $\sum_{k=1}^{m}\sum_{l=1}^{d} \dfrac{\partial f(X)^j_i}{\partial X_l^k}\bigg|_{X_0}$ as B, for conciseness):
    \begin{align}
        &\int_{X^0 - [\delta/2]_{m,d}}^{X^0 + [\delta/2]_{m,d}} \big|B \cdot (X_l^k - X_l^{k, 0}) \big|^p dX \nonumber \\\nonumber\\
        = \int_{X_l^{k, 0} - \delta/2}^{X_l^{k, 0} + \delta/2}& \big|B \cdot (X_l^k - X_l^{k, 0}) \big|^p dX_l^k \int_{X^{-k, 0}_{-l} - [\delta/2]_{m,d, -l}^{-k}}^{X^{-k, 0}_{-l} + [\delta/2]_{m,d, -l}^{-k}}dX_{-i}^{-k} \nonumber
    \end{align}
    $X_{-i}^{-k}$ denotes the elements of $X$ other than the $i$-th component of the $k$-th sequence item. Note that the integral is truly a multiple integral over all the dimensions, we write it in a short just for conciseness. The second integral represents $md - 1$ integrals over the constant $1$, each over a range of $\delta$. Thus, this value is $\Big(\delta^{md-1}\Big)$. We can compute the first integral as:
    \begin{align}
        &\int_{X_l^{k, 0} - \delta/2}^{X_l^{k, 0} + \delta/2} \big|B \cdot (X_l^k - X_l^{k, 0}) \big|^p dX_l^k \nonumber \\\nonumber \\ =&\int_{X_l^{k, 0} - \delta/2}^{X_l^{k, 0} + \delta/2} |B|^p \cdot |X_l^k - X_l^{k, 0}| \> dX_l^k \nonumber \\\nonumber \\
        =&\>|B|^p \int_{-\delta/2}^{\delta/2} |u|^p du \quad\text{[$u = X_l^k - X_l^{k, 0}$]} \nonumber \\\nonumber \\
        =&\>|B|^p \cdot 2 \cdot \int_{0}^{\delta/2} u^p du \nonumber \\\nonumber \\
        =&\>2 \cdot |B|^p \cdot \dfrac{u^{p+1}}{p+1}\bigg|_0^{\delta/2}  \nonumber\\\nonumber \\
        =&\>\Bigg|\sum_{k=1}^{m}\sum_{l=1}^{d} \dfrac{\partial f(X)^j_i}{\partial X_l^k}\bigg|_{X_0}\Bigg|^p \cdot \dfrac{1}{p + 1} \cdot 2^{-p} \cdot \delta^{p + 1} \nonumber
    \end{align}
    Thus, we have,
    \begin{align}
        &\int_{X^0 - [\delta/2]_{m,d}}^{X^0 + [\delta/2]_{m,d}} \sum_{i=1}^{d}\sum_{j=1}^{n}\big|\overline{f}(X)_i^j - f(X)_i^j\big|^p dX \nonumber \\\nonumber\\
        =& \sum_{i=1}^{d}\sum_{j=1}^{n} \Bigg|\sum_{k=1}^{m}\sum_{l=1}^{d} \dfrac{\partial f(X)^j_i}{\partial X_l^k}\bigg|_{X_0}\Bigg|^p \cdot \dfrac{1}{p + 1} \cdot 2^{-p} \cdot \delta^{p + 1} \cdot\delta^{md-1} \nonumber \\\nonumber\\
        =& \sum_{i=1}^{d}\sum_{j=1}^{n} \Bigg|\sum_{k=1}^{m}\sum_{l=1}^{d} \dfrac{\partial f(X)^j_i}{\partial X_l^k}\bigg|_{X_0}\Bigg|^p \cdot \dfrac{1}{p + 1} \cdot 2^{-p} \cdot \delta^{p + md} \nonumber
    \end{align}
    We consider a covering $\mathcal{X}^0$ of $\mathcal{S}$, such that $[X^0 - [\delta/2]_d, X^0 + [\delta/2]_d) \in \mathcal{X}^0$, we have:
    \begin{align}
        \bigg( \int_\mathcal{S} \sum_{i=1}^{d}\sum_{j=1}^{n}\big|\overline{f}(X)_i^j - f(X)_i^j\big|^p dX \bigg)^\mathlarger{\frac{1}{p}} &\leq \epsilon \nonumber \\\nonumber\\
        \Rightarrow\sum_{X^0 \in \mathcal{X}^0} \int_{X^0 - [\delta/2]_{m,d}}^{X^0 + [\delta/2]_{m,d}} \sum_{i=1}^{d}\sum_{j=1}^{n}\big|\overline{f}(X)_i^j - f(X)_i^j\big|^p dX &\leq \epsilon^p \nonumber \\\nonumber\\
        \Rightarrow\sum_{X^0 \in \mathcal{X}^0}  \sum_{i=1}^{d}\sum_{j=1}^{n} \Bigg|\sum_{k=1}^{m}\sum_{l=1}^{d} \dfrac{\partial f(X)^j_i}{\partial X_l^k}\bigg|_{X_0}\Bigg|^p \cdot \dfrac{1}{p + 1} \cdot 2^{-p} \cdot \delta^{p + md} &\leq \epsilon^p \nonumber \\\nonumber\\
        \Rightarrow\dfrac{\delta^{p+md}}{\displaystyle 2^p \cdot (p+1)}  \Bigg(\sum_{X^0 \in \mathcal{X}^0} \sum_{i=1}^{d}\sum_{j=1}^{n} \Bigg|\sum_{k=1}^{m}\sum_{l=1}^{d} \dfrac{\partial f(X)^j_i}{\partial X_l^k}\bigg|_{X_0}\Bigg|^p \Bigg) &\leq \epsilon^p \nonumber \\\nonumber\\
        \Rightarrow\delta \leq \Bigg(\dfrac{2^p \cdot (p+1) \cdot \epsilon^p}{\sum_{X^0 \in \mathcal{X}^0} \sum_{i=1}^{d}\sum_{j=1}^{n} \Big|\sum_{k=1}^{m}\sum_{l=1}^{d} \frac{\partial f(X)^j_i}{\partial X_l^k}\big|_{X_0}\Big|^p}&\Bigg)^{\mathlarger{\frac{1}{(p+md)}}} \nonumber
    \end{align}
\end{proof}
\section{Additional Experiments}\label{asec:additional-expts}

\begin{figure}[h!]
    \centering
    \includegraphics[width=0.45\textwidth]{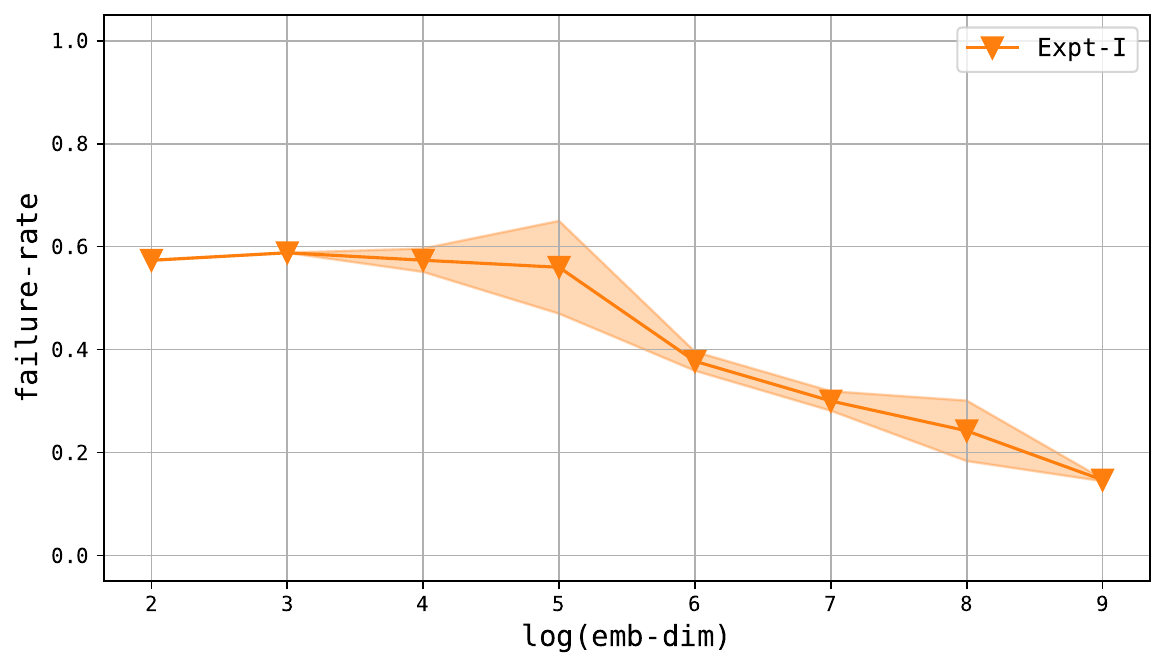}
    \caption{Trend of {\tt failure-rate} with respect to embedding dimension for $20$ classes in \textsc{Expt-II}. We keep $r=d=128$ for this experiment, found to be the best configuration for \textsc{Expt-II} from Figure \ref{fig:trend-dims}.}
    \label{fig:trend-classes}
\end{figure}

\textbf{Trend with \#classes}. To check the effect of higher number of classes in \textsc{Expt-II}, we repeat the trend analysis study (similar to Figure \ref{fig:trend-dims-emb}) with $20$ classes, Figure \ref{fig:trend-classes}. We posited earlier that for piecewise constant functions, small-sized pieces would adversely affect performance. The number of pieces would increase as the number of classes increases. With the range constant, this implies that a higher number of classes would lead to smaller pieces. Hence, we expect a higher {\tt failure-rate} as the number of classes increases. However, as the embedding dimension increases, the positioning of the constant pieces becomes sparse, if the number of classes is kept constant. Hence, we expect {\tt failure-rate} to reduce as the number of embedding dimensions is increased. That is exactly what we observe in the figure.

\begin{figure*}
    \centering
    \begin{subfigure}[b]{0.33\textwidth}
        \includegraphics[width=\linewidth]{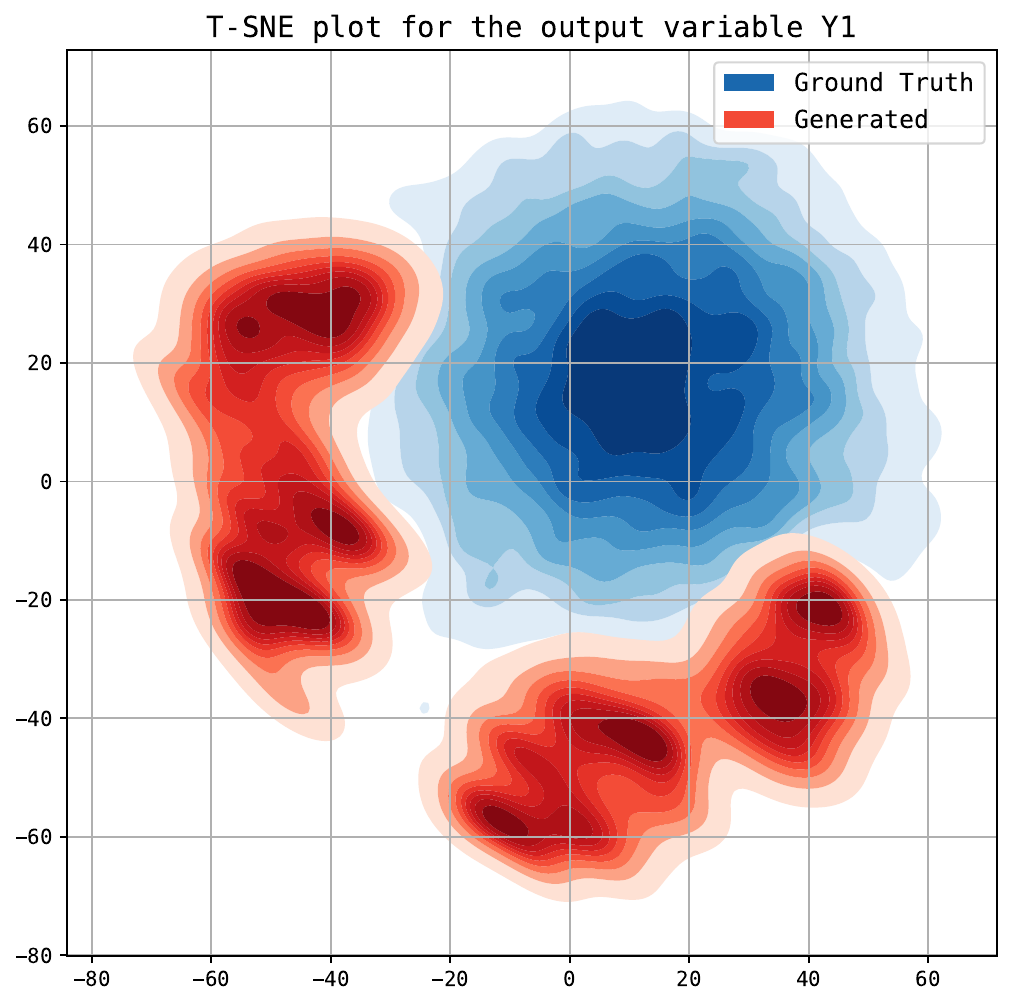}
        \caption{}
        \label{fig:t-sne-y1}
    \end{subfigure}%
    \begin{subfigure}[b]{0.33\textwidth}
        \includegraphics[width=\linewidth]{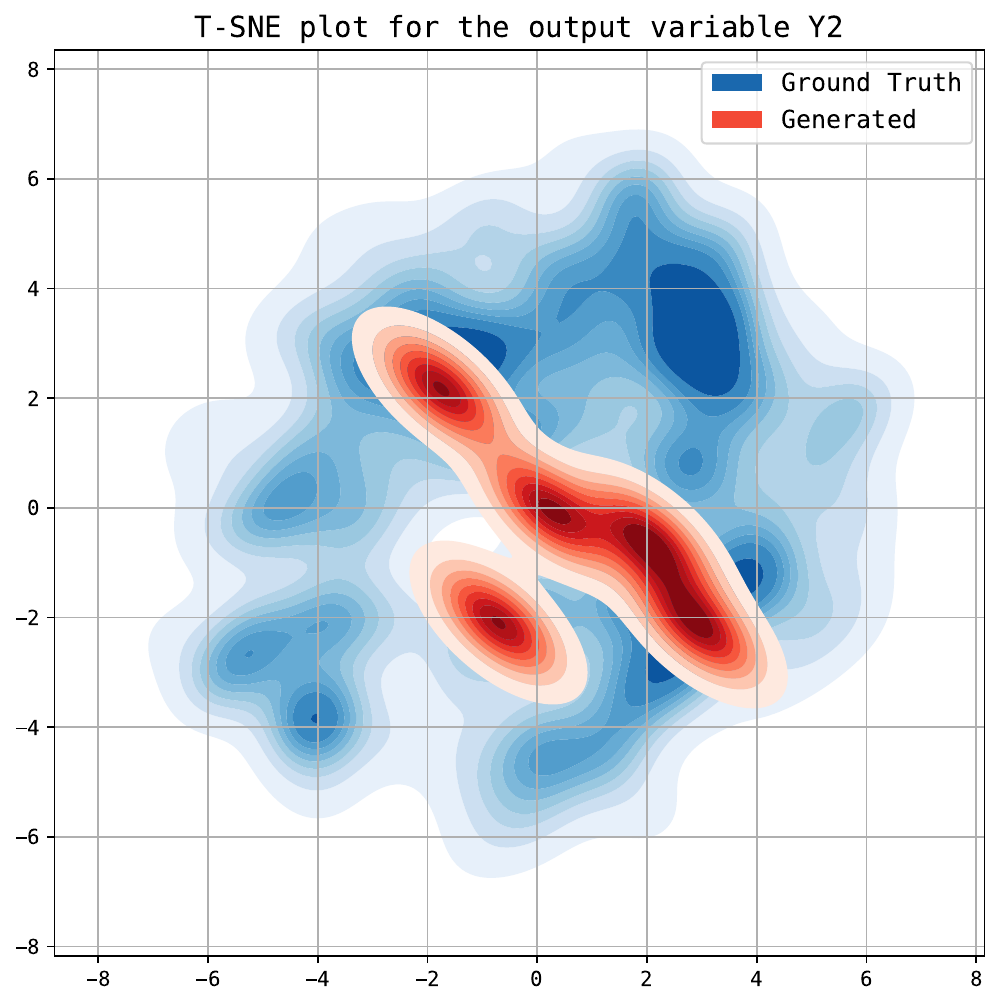}
        \caption{}
        \label{fig:t-sne-y2}
    \end{subfigure}%
    \begin{subfigure}[b]{0.33\textwidth}
        \includegraphics[width=\linewidth]{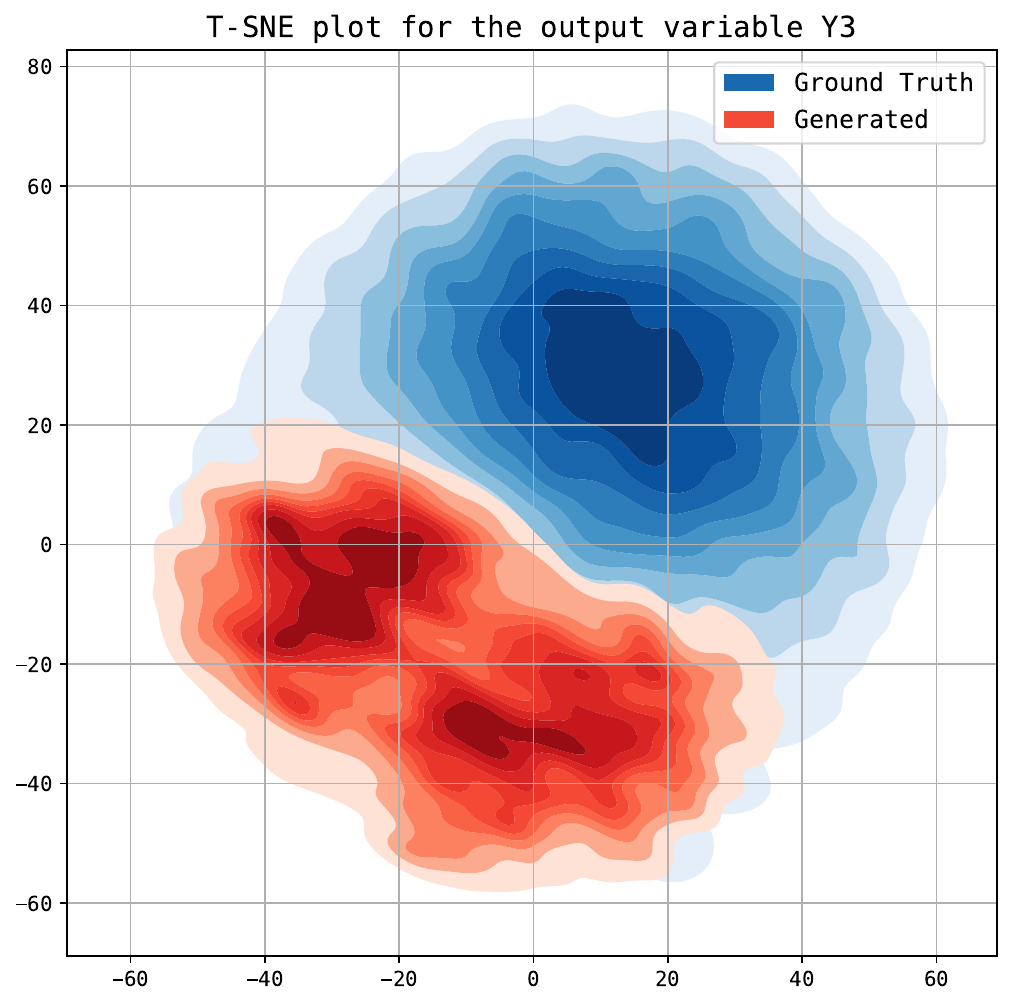}
        \caption{}
        \label{fig:t-sne-y3}
    \end{subfigure}
    \caption{\textsc{t-sne} plot for the generated and ground truth $Y_1$ (a), $Y_2$ (b) and $Y_3$ (c). Deeper colors indicate higher density.}
    \label{fig:t-sne-y*}
\end{figure*}

We also qualitatively analyze the generations in \textsc{Expt-I} to visualize the failure of the Transformer in generating $Y_1$, $Y_2$, and $Y_3$. We include \textsc{t-sne} plots of the generated and ground truth values for these three variables, Figures \ref{fig:t-sne-y1}, \ref{fig:t-sne-y2} and \ref{fig:t-sne-y3}. For all of the variables, we see that the Transformer fails to generate representative vectors -- the densities of generated and ground truth points differ significantly.

\begin{figure}[h!]
    \centering
    \includegraphics[width=0.40\textwidth]{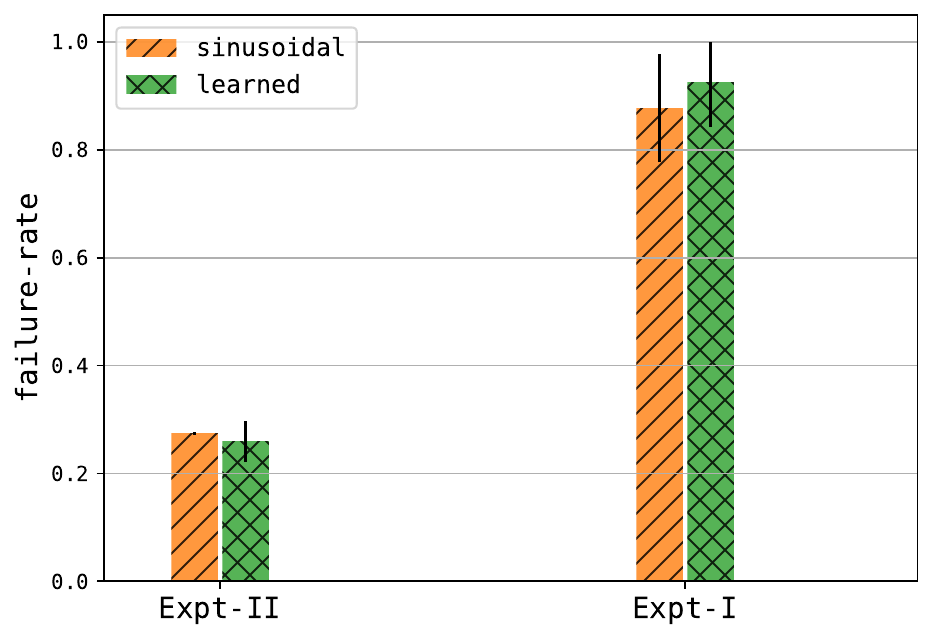}
    \caption{Trend of {\tt failure-rate} with respect to two popular Positional Embedding schemes: sinusoidal \citep{vaswani-etal-2017-transformer} and learned \citep{devlin-etal-2019-bert}. We see that changing the scheme does not affect the performance significantly. For \textsc{Expt-I}, we keep $r=d=32$, and for \textsc{Expt-II}, we keep $r=d=128$. These configurations were found to be best performing from Figure \ref{fig:trend-dims}.}
    \label{fig:trend-pe-scheme}
\end{figure}

We also examine the performance across two popular positional embedding schemes: sinusoidal \citep{vaswani-etal-2017-transformer}, learned \citep{devlin-etal-2019-bert}, Figure \ref{fig:trend-pe-scheme}. We observe a similar thing -- there is a significant difference between performance in \textsc{Expt-I} and \textsc{Expt-II} irrespective of the positional embedding scheme. These experiments answer all the questions we set in Section \ref{subsec:synth-data-designs}. We test for a few more things too: (\textit{a}) checking performance in \textsc{Expt-II} for a higher number of classes, and (\textit{b}) checking performance as $k$ increases in {\tt failure-rate@}$k$.

\begin{figure}[h!]
    \centering
    \includegraphics[width=0.40\textwidth]{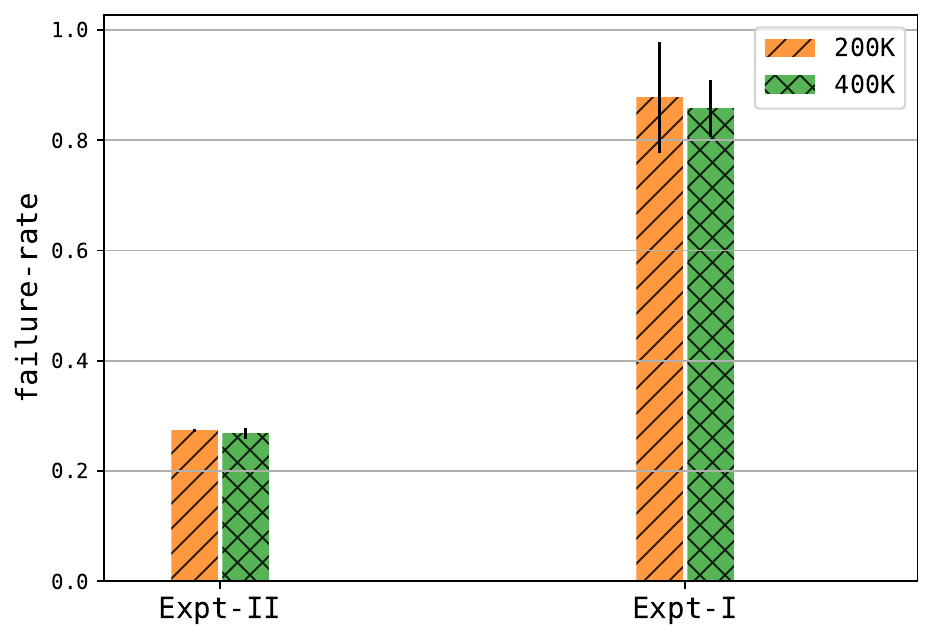}
    \caption{Effect of training dataset size on the performance. {\tt 200K} signifies dataset with $200,000$ training samples, while {\tt 400K} signifies dataset with $400,000$ training samples. We see that doubling the dataset size does not affect the performance significantly. For \textsc{Expt-I}, we keep $r=d=32$, and for \textsc{Expt-II}, we keep $r=d=128$. These configurations were found to be best performing from Figure \ref{fig:trend-dims}.}
    \label{fig:trend-data-size}
\end{figure}

We examine how the performance changes on changing the dataset size (Figure \ref{fig:trend-data-size}). We find that doubling the dataset has very little effect on the performance for \textsc{Expt-I} and \textsc{Expt-II}. This helps us eliminate any effect of data size on the conclusions we draw from all the experiments.

\section{Training Details}\label{asec:training-details}
We perform an extensive hyperparameter search for each experimental setting using \href{https://wandb.ai}{\tt wandb.ai}. We search for the best combination for the following hyperparameters: {\it batch size}, {\it maximum number of training steps}, {\it learning rate}, {\it number of attention heads}, {\it number of Transformer layers}, {\it dropout} and {\it warmup steps for learning rate}. We list the range of each in Table \ref{tab:hyperparam-search-space}. We found that the best combination for each setup is not the same.

We train all our models on a single A$100$ GPU, with each training consuming $\sim 2.5$ GB GPU memory and $\sim 1-3$ hours of training.

\begin{table}[h]
    \centering
    \caption{Search Space for Hyperparamters. \{$\cdot$\} denotes a set, $\sim \mathcal{U}(l, h)$ denotes sampling from a Uniform distribution within $l$ and $h$. {\tt num-layers} corresponds to the \textit{number of layers} in the Transformer, {\tt max-steps} corresponds to the \textit{maximum training steps}, {\tt num-attn-heads} corresponds to the \textit{number of attention heads}, {\tt warmup-steps} corresponds to the fraction of training steps used for \textit{learning rate warmup}, and {\tt emb-dim} denotes the embedding dimension of the Transformer.\\}
    \begin{tabular}{*2c}
        \toprule
        \textbf{Hyperparameter} & \textbf{Values} \\
        \midrule
        {\tt batch-size} & \{$128$, $256$, $512$\} \\
        {\tt max-steps} & \{$1200$, $1400$, $1600$\} \\
        {\tt num-layers} & \{$2$, $4$, $6$\} \\
        {\tt num-attn-heads} & \{$2$, $4$, $\cdots$, $\min$($16$, {\tt emb-dim})\} \\
        {\tt learning-rate} & $\sim \mathcal{U}(10^{-2}, 5 \times 10^{-6})$ \\
        {\tt warmup-steps} & $\sim \mathcal{U}(0.2, 0.4)$ \\
        {\tt dropout} & $\sim \mathcal{U}(0.1, 0.2)$ \\
        \bottomrule
    \end{tabular}
    \label{tab:hyperparam-search-space}
\end{table}

\section{Details on Auxiliary Datasets}\label{asec:aux-datasets}

We had provided ablation studies on how the performance varies for different number of inputs and outputs. Here, we provide the equations governing the generation of those datasets. These generation equations are minor variations of equations in Table \ref{tab:data-gen-equations}, which generates data for $4$ inputs and $3$ outputs. Tables \ref{tab:data-gen-equations-m2n3}, \ref{tab:data-gen-equations-m3n3}, \ref{tab:data-gen-equations-m4n1}, \ref{tab:data-gen-equations-m4n2} show the equations for ablations on $2$ inputs, $3$ inputs, $1$ output and $2$ outputs.

\begin{table}[h]
    \centering
    \caption{Equations governing the synthetic data generation. \textit{Type: Input} signifies that these variables are fed to the Transformer Encoder. \textit{Type: Output} signifies that the Transformer is trained to generate these variables through the Decoder. By design, we have functions covering the whole spectrum of first-order derivatives ($>1$, $=1$ and $<1$).\\}
    
    \begin{tabular}{cc}
        \toprule
        \textbf{Type} & \textbf{Generator Function} \\
        \midrule
        \multirow{5}{*}{Input} & $X_1 \sim Uniform(-1, 1)$ \\[0.50em] 
        & $X_2 = \sqrt[3]{X_1}$\\[0.50em]
        & $X_3$ = $2 \cdot \log(2 + X_1) + X_2 / 10$ \\[0.50em]
        & $X_4$ = $e^{X_2} + X_3$ \\[0.50em]
        \midrule
        \multirow{4}{*}{Output} & $Y_1$ = $\dfrac{1}{5}(X_1 + X_2 + X_3 + X_4)$ \\[0.50em]
        & $Y_2$ = $X_1 \cdot Y_1 + e^{X_2} + X_3 + \log(X_4)$ \\[0.50em]
        & $Y_3$ = $\dfrac{1}{5}(X_1 + Y_2 + Y_1 + \sqrt{X_2} + X_3 \cdot X_4)$\\[0.50em]
        \bottomrule
    \end{tabular}
    \label{tab:data-gen-equations}
\end{table}

\begin{table}[h]
    \centering
    \caption{Equations governing the synthetic data generation for the auxiliary datasets.}
    \begin{subtable}[h]{0.45\textwidth}
        \centering
        \caption{Equations governing the synthetic data generation for $2$ inputs and $3$ outputs.}
        \begin{tabular}{cc}
            \toprule
            \textbf{Type} & \textbf{Generator Function} \\
            \midrule
            \multirow{3}{*}{Input} & $X_1 \sim Uniform(-1, 1)$ \\[0.50em] 
            & $X_2 = \sqrt[3]{X_1}$\\[0.50em]
            \midrule
            \multirow{4}{*}{Output} & $Y_1$ = $\dfrac{1}{5}(X_1 + X_2)$ \\[0.50em]
            & $Y_2$ = $X_1 \cdot Y_1 + e^{X_2}$ \\[0.50em]
            & $Y_3$ = $\dfrac{1}{5}(X_1 + Y_2 + Y_1 + \sqrt{X_2})$\\[0.50em]
            \bottomrule
        \end{tabular}
        \label{tab:data-gen-equations-m2n3}
    \end{subtable}
    \hfill
    \begin{subtable}[h]{0.45\textwidth}
        \centering
        \caption{Equations governing the synthetic data generation for $3$ inputs and $3$ outputs.}
        \begin{tabular}{cc}
            \toprule
            \textbf{Type} & \textbf{Generator Function} \\
            \midrule
            \multirow{4}{*}{Input} & $X_1 \sim Uniform(-1, 1)$ \\[0.50em] 
            & $X_2 = \sqrt[3]{X_1}$\\[0.50em]
            & $X_3$ = $2 \cdot \log(2 + X_1) + X_2 / 10$ \\[0.50em]
            \midrule
            \multirow{4}{*}{Output} & $Y_1$ = $\dfrac{1}{5}(X_1 + X_2 + X_3)$ \\[0.50em]
            & $Y_2$ = $X_1 \cdot Y_1 + e^{X_2} + X_3$ \\[0.50em]
            & $Y_3$ = $\dfrac{1}{5}(X_1 + Y_2 + Y_1 + \sqrt{X_2})$\\[0.50em]
            \bottomrule
        \end{tabular}
        \label{tab:data-gen-equations-m3n3}
     \end{subtable}%

    \begin{subtable}[h]{0.45\textwidth}
        \centering
        \caption{Equations governing the synthetic data generation for $4$ inputs and $1$ output.}
        \begin{tabular}{cc}
            \toprule
            \textbf{Type} & \textbf{Generator Function} \\
            \midrule
            \multirow{5}{*}{Input} & $X_1 \sim Uniform(-1, 1)$ \\[0.50em] 
            & $X_2 = \sqrt[3]{X_1}$\\[0.50em]
            & $X_3$ = $2 \cdot \log(2 + X_1) + X_2 / 10$ \\[0.50em]
            & $X_4$ = $e^{X_2} + X_3$ \\[0.50em]
            \midrule
            \multirow{1}{*}{Output} & $Y_1$ = $\dfrac{1}{5}(X_1 + X_2 + X_3 + X_4)$ \\[0.50em]
            \bottomrule
        \end{tabular}
        \label{tab:data-gen-equations-m4n1}
    \end{subtable}
    \hfill
    \begin{subtable}[h]{0.45\textwidth}
        \centering
        \caption{Equations governing the synthetic data generation for $4$ inputs and $2$ outputs.}
        \begin{tabular}{cc}
            \toprule
            \textbf{Type} & \textbf{Generator Function} \\
            \midrule
            \multirow{5}{*}{Input} & $X_1 \sim Uniform(-1, 1)$ \\[0.50em] 
            & $X_2 = \sqrt[3]{X_1}$\\[0.50em]
            & $X_3$ = $2 \cdot \log(2 + X_1) + X_2 / 10$ \\[0.50em]
            & $X_4$ = $e^{X_2} + X_3$ \\[0.50em]
            \midrule
            \multirow{2}{*}{Output} & $Y_1$ = $\dfrac{1}{5}(X_1 + X_2 + X_3 + X_4)$ \\[0.50em]
            & $Y_2$ = $X_1 \cdot Y_1 + e^{X_2} + X_3 + \log(X_4)$ \\[0.25em]
            \bottomrule
        \end{tabular}
        \label{tab:data-gen-equations-m4n2}
    \end{subtable}
    \label{tab:data-gen-equations-aux}
\end{table}
\clearpage

\end{document}